\documentclass{article}

\usepackage[nonatbib, final]{neurips_2025}

\usepackage[utf8]{inputenc} 
\usepackage[T1]{fontenc}    
\usepackage{hyperref}       
\usepackage{url}            
\usepackage{booktabs}       
\usepackage{amsfonts}       
\usepackage{nicefrac}       
\usepackage{microtype}      
\usepackage{xcolor}         

\usepackage{amsmath,amsfonts,bm}

\def\eqref#1{equation~\ref{#1}}

\def\1{\bm{1}}

\DeclareMathAlphabet{\mathsfit}{\encodingdefault}{\sfdefault}{m}{sl}
\SetMathAlphabet{\mathsfit}{bold}{\encodingdefault}{\sfdefault}{bx}{n}

\DeclareMathOperator*{\argmax}{arg\,max}
\DeclareMathOperator*{\argmin}{arg\,min}

\newcommand{\btheta}{{\boldsymbol{\theta}}}

\newcommand{\method}{PoE-World }
\newcommand{\methodwospace}{PoE-World}
\usepackage{url}
\usepackage{booktabs}
\usepackage{amsmath,amsthm,amsfonts,amssymb,amscd}
\usepackage{bbm}
\usepackage{cleveref}
\usepackage{lastpage}
\usepackage{enumerate}
\usepackage{enumitem}
\usepackage{fancyhdr}
\usepackage{mathrsfs}
\usepackage{xcolor}
\usepackage{graphicx}
\usepackage{listings}
\usepackage{mathtools}
\usepackage{todonotes}
\usepackage{adjustbox}
\usepackage{wrapfig}

\usepackage{multirow}
\usepackage{array}
\usepackage{dblfloatfix}
\usepackage{makecell}
\usepackage{spverbatim}
\usepackage{appendix}
\usepackage{bbm}
\usepackage{comment}
\usepackage{subcaption}
\usepackage{alltt}

\title{PoE-World: Compositional World Modeling with \\Products of Programmatic Experts}

\author{%
  \textbf{Wasu Top Piriyakulkij}$^1$ \qquad 
  \textbf{Yichao Liang}$^2$ \qquad 
  \textbf{Hao Tang}$^1$ \\
  \textbf{Adrian Weller}$^{2,3}$ \qquad 
  \textbf{Marta Kryven}$^4$ \qquad 
  \textbf{Kevin Ellis}$^1$\\
  Cornell University$^1$ \quad University of Cambridge$^2$ \\
  The Alan Turing Institute$^3$ \qquad Dalhousie University$^4$
}

\begin{document}

\maketitle

\begin{abstract}
Learning how the world works is central to building AI agents that can adapt to complex environments. 
Traditional world models based on deep learning demand vast amounts of training data, and do not flexibly update their knowledge from sparse observations. 
Recent advances in program synthesis using Large Language Models (LLMs) give an alternate approach which learns world models represented as source code, supporting strong generalization from little data. 
To date, application of program-structured world models remains limited to natural language and grid-world domains. We introduce a novel program synthesis method for effectively modeling complex, non-gridworld domains by representing a world model as an exponentially-weighted product of programmatic experts (\methodwospace) synthesized by LLMs.
We show that this approach can learn complex, stochastic world models from just a few observations. 
We evaluate the learned world models by embedding them in a model-based planning agent, demonstrating efficient performance and generalization to unseen levels on Atari's Pong and Montezuma's Revenge. 
We release our code and display the learned world models and videos of the agent's gameplay at \url{https://topwasu.github.io/poe-world}.
\end{abstract}

\section{Introduction}

\begin{figure*}[t]
\centering
\includegraphics[width=0.93\linewidth]{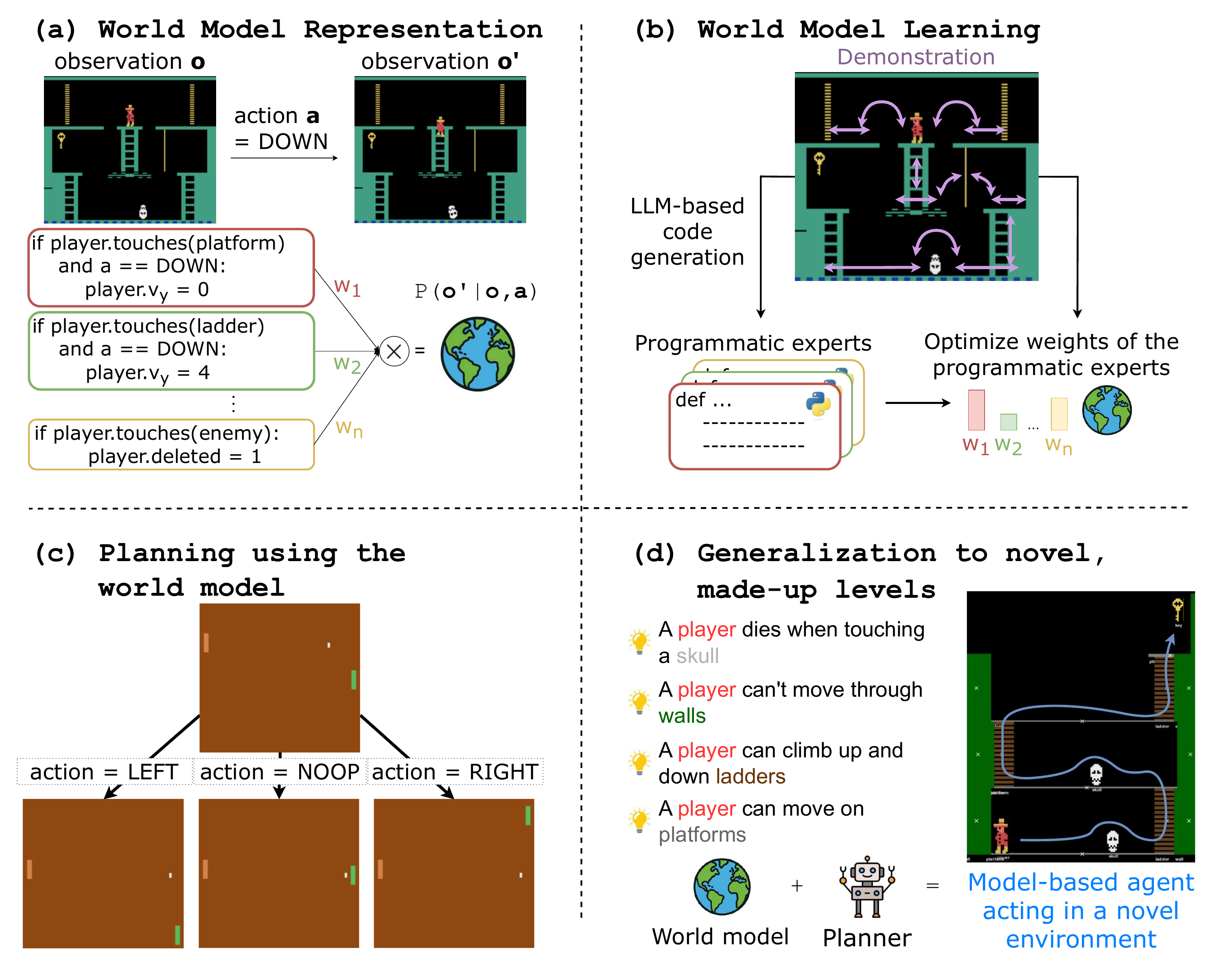}
\caption{(a) World models predict the next state given a state-action history.
We do this with a product of experts of many small programs.
(b) The learner is given a short (<1 min) demonstration of gameplay as input, and uses it to synthesize initial world-model programs. These programs are refined online in later environment interactions.
(c) World models support planning by imagining future states.
(d) Symbolic programs encode abstract knowledge that generalizes to new game levels.}
\label{fig:overview}
\end{figure*}

How should an intelligent agent represent the dynamics of the natural world?
We want a representation that is efficiently learnable, yet flexible enough to handle stochasticity and partial observability, and which supports planning and decision-making.
Neural network world models such as Dreamer~\cite{hafner2023mastering} are radically flexible, but demand enormous training data (compared to humans \cite{tsividis2017human}).
Symbolic world models such as WorldCoder~\cite{tang2024worldcoder} instead generate a Python program to represent how the world works.
These programmatic world models are data-efficient, because program synthesis requires less data than neural network training---
but struggle to scale beyond simple gridworlds, as they do a discrete combinatorial search to find a single large program describing everything about how the world works.

We take inspiration from a longstanding view in philosophy and cognitive science of the mind as a community of interacting experts~\cite{fodor1983modularity,minsky1986society, dennett1993consciousness,marcus2003algebraic}. This modular organization is evident across multiple scales in natural intelligence, from functional specialization of brain systems \cite{gazzanigacognitive} to the distinct learning trajectories of specific skills \cite{anderson1982acquisition,liu2024automated}. We integrate this modular perspective with the computational paradigm of learning as Program Synthesis, 
which models learned concepts as symbolic programs \cite{lake2015human,lake2020people,tian2020learning,hofer2021learning,rule2024symbolic,liang2022drawing}.
We extend this line of work by proposing a new computational account of learning world models: as the acquisition of context-specific expert programs, which are refined through practice and reused compositionally to support flexible, goal-directed behavior.

Algorithmically, our key idea is to  \emph{decompose the problem of learning a world program into learning hundreds of small programs}. Each of these learned programs encodes a different causal law, which we probabilistically aggregate to predict future observations (\Cref{fig:overview}a).
This makes our world knowledge more modular, and also more learnable, because we no longer search for a single monolithic program handling everything at once.
The resulting system, which we call \method (\textbf{P}roduct~\textbf{o}f~programmatic~\textbf{E}xperts), can build elaborate world models 
that accurately support planning and reinforcement learning (RL) from even a brief demonstration in complex Atari games, such as Montezuma's Revenge. 
\method handles stochasticity because the product of programs is probabilistic, and, as we show, further handles partial observability.
To the best of our knowledge, this is the first time a symbolic world model has been learned for environments of this complexity.

Importantly, although \method models fine-grained pixel-level movement, it does \emph{not} model pixel-level visual appearance, instead assuming symbolic observations from an object detector.
Unlike model-based reinforcement learning, \method does not attempt efficient exploration, but focuses  on faithfully learning from a demonstrated trajectory
(\Cref{fig:overview}b).
Lastly, while the world models are ultimately used for planning (\Cref{fig:overview}c), \method fundamentally focuses on world modeling, and not on solving the challenging computational problem of planning itself.

Despite these limitations, we view \method as addressing a central learning problem:
Given limited demonstrations of a new environment, quickly assemble a working world model that compositionally generalizes to new situations (\Cref{fig:overview}d).
We highlight the following contributions:
\begin{enumerate}[leftmargin=*]
    \item The \method representation and learning algorithm for symbolic world models.
    \item An empirical study of \method on two representative Atari games, Pong and Montezuma's Revenge, demonstrating its superior learning efficiency compared to deep RL, and improved scalability compared to state-of-the-art symbolic model-based RL: \method can synthesize 4000+ line programs that generalize zero-shot to novel game levels and game variations.
    \item Demonstration of how to use \methodwospace's world models for planning-based decision making, and as a simulated pretraining environment for deep RL.
\end{enumerate}

\section{Background: World Model Learning}

A sequential decision-making problem can be described as $(\mathcal{O}, \mathcal{A}, P, R)$ where $\mathcal{O}$ is an observation space, $\mathcal{A}$ is an action space, $P$ is an environment dynamics $P = p_{env}(o_{t+1}|o_{1:t}, a_{1:t})$ where $o \in \mathcal{O}$ and $a \in \mathcal{A}$, and $R$ is a reward function. 
This full-history environment formulation is mathematically equivalent to a Partially Observable Markov Decision Process (POMDP) formulation \cite{aastrom1965optimal}. 
In the setting where $P$ is unknown to an agent, the agent learns by interacting with the environment and observing transitions $(o_t, a_t, o_{t+1}, r_t)$ at timestep $t$. 


We focus on 
learning the world model $\hat{P} = p_{model}(o_{t+1}|o_{1:t}, a_{1:t})$ which approximates the true, unknown environment dynamics $P = p_{env}(o_{t+1}|o_{1:t}, a_{1:t})$ given observed trajectories $D = \{\tau_i\}_{i=1}^n$ where each trajectory is a sequence of observations and actions $\tau = (o_{1:T+1}, a_{1:T})$ for some $T$ ($r_{1:T}$ is also a part of a trajectory, but we drop it for simplicity). 
The learned world model will later be used by a model-based agent, either through lookahead planning or RL training, to act in the environment.

We treat the world modeling problem as an optimization problem, specifically empirical risk minimization:
\begin{equation}
    p_{model}^* = \argmin_{p_{model}} \sum_{ (o_{1:T+1}, a_{1:T})\in D} \sum_{t =1}^T \ell (p_{model};o_{1:t+1}, a_{1:t})
    \label{eq:erm}
\end{equation}
where $\ell$ is a loss function, such as the negative log likelihood function $\ell (p_{model};o_{1:t+1}, a_{1:t}) = -\log \; p_{model}(o_{t+1}|o_{1:t}, a_{1:t})$.

Previous works in the model-based reinforcement learning literature \cite{ha2018world, schrittwieser2020mastering, hafner2023mastering, kaiser2020atari100k, iris2023, alonso2024diamond, zhou2024diffusion} have used various deep neural network architectures, including convolutional and recurrent neural networks, transformers, and diffusion models, to define $p_{model}$ as a parametric model $p_{model} = p_\btheta$. 
Then, $\btheta$ is optimized via gradient descent. A weakness of such approaches is poor sample efficiency and generalization. 
For example, Diamond~\cite{alonso2024diamond} has failure modes  such as imagining that the player can walk through a wall or teleport, 
even after training on almost 100 hours of observations. 

Other works leverage Large Language Models (LLMs) to synthesize a code world model \cite{tang2024worldcoder, dainese2024codeworldmodel, ahmed2025synthesizing}, which can be seen as searching for programmatic $p_{model}$.
These LLM-based code generation algorithms are more sample efficient and extrapolate more systematically than  deep learning approaches.
However, they have yet to succeed beyond  simple text-based and gridworld games. 

\section{Modeling the World as a Product of Programmatic Experts (\methodwospace)}

To address the limitations of existing world model learning methods, we propose representing world models as exponentially-weighted \textbf{P}roducts \textbf{o}f programmatic \textbf{E}xperts (\textbf{\methodwospace}), enabling sample-efficient and scalable learning of probabilistic world models that leverages LLM code generation. 
\Cref{fig:overview} visualizes both the representation and learning algorithm discussed in this section. 

\subsection{World Model Representation: Product of Programmatic Experts (\methodwospace)}\label{sec:rep}

\begin{figure*}[t]
\centering
\includegraphics[width=0.8\linewidth]{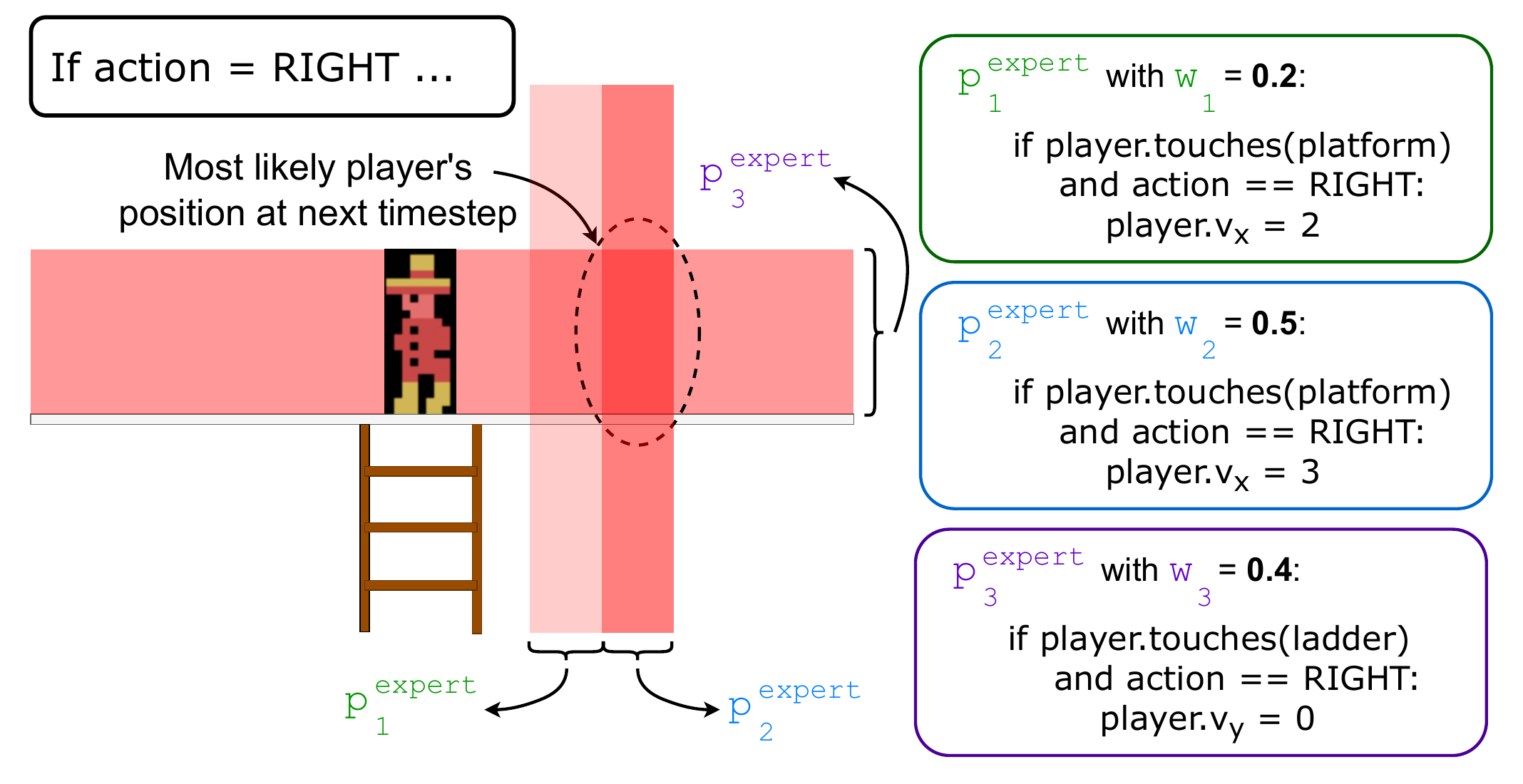}
\vspace{-1pt}
\caption{A ``heat map'' illustration of how simple Python programs are interpreted as distributions, and how they are combined into a single distribution over next-timestep object locations.}
\label{fig:heatmap}
\end{figure*}

World models represented as exponentially weighted \textbf{P}roducts \textbf{o}f programmatic \textbf{E}xperts (\textbf{\methodwospace}) can be described mathematically as follows:
\begin{equation}
    p_\btheta(o_{t+1}|o_{1:t}, a_{1:t}) \propto \prod_i p^{expert}_i(o_{t+1}|o_{1:t}, a_{1:t})^{\theta_i}
    \label{eq:rep}
\end{equation}
where $p^{expert}_i$ are programs and $\theta_i$ are their associated scalar weights. 

This representation enables modularity and compositionality. It allows composing many small programs into a full world model to capture complicated environmental dynamics (see \Cref{fig:heatmap}). 
We can think of each program as an expert which expresses opinions about particular aspects of the world. 
For example, in the context of modeling a video game environment, one expert might encode ``if a player touches a skull, then the player dies,'' while another encodes ``if an action is LEFT when the player is on a platform, the player's x-axis velocity is -2.'' 
The former expert does not express opinions on the player's movement, while the latter does not mention conditions for player's death. 

\paragraph{A factored state representation yields tractable inference.}
We assume an object-centric state where each object has a bounding box and velocities, each stored as a separate slot, attribute, or feature.
We treat each such feature as conditionally independent, given the history so far.
This independence assumption makes \Cref{eq:rep} tractable, because we can compute a separate normalizing constant for each feature.
Concretely, let $f$ index the different object features of the next observation $o_{t+1}$, and let $o^f$ be the indexing of $o$ with feature $f$.
Then,
\begin{equation*}
    p_\btheta(o_{t+1}|o_{1:t}, a_{1:t}) = \prod_f\frac{1}{Z_f}\prod_i p^{expert}_i(o_{t+1}^f|o_{1:t}, a_{1:t})^{\theta_i}\quad Z_f=\sum_{o^f}\prod_i p^{expert}_i(o^f|o_{1:t}, a_{1:t})^{\theta_i}
\end{equation*}

\paragraph{Benefits of the full-history formulation over POMDP formulation.} The full-history environment formulation is formally equivalent to a POMDP which instead compresses the history into a Markov latent state.
We use the history formulation because it makes our world models more modular.
Learning global latent variables would entangle all the experts, because every expert would condition on the latent state.
Therefore, learning a new latent variable (e.g. ``how long the player has been falling'') changes the input/output space of every expert, necessitating global joint updates to the structure of every program (e.g. an expert for ``is the player dead'' would need to be updated).
The history formulation allows independent learning of independent mechanisms.


\paragraph{Hard constraints.}
Atari (and the real world) is too complex to perfectly simulate with any effectively learnable program.
Therefore, our probabilistic model tends to over-approximate the set of possible futures, giving fuzzy approximate predictions.
For example, in Montezuma's Revenge, instead of perfectly modeling the physics of falling downward and landing on the ground, we predict a generic downward trajectory.
Ideally that trajectory would perfectly enforce the constraint that the player lands flat on the ground, and never sinks into the ground, but a fuzzy stochastic expert for falling downward could violate that constraint.
Therefore, to sharpen the model's outputs, we further learn a collection of hard constraints, $\left\{ c_j \right\}$, where $c_j: \mathcal{O} \rightarrow \{0, 1\}$:
\begin{equation}
    p_\btheta(o_{t+1}|o_{1:t}, a_{1:t}) \propto \prod_i p^{expert}_i(o_{t+1}|o_{1:t}, a_{1:t})^{\theta_i} \cdot \mathbbm{1}\left[ \bigvee_j c_j(o_{t+1})  \right]
    \label{eq:rep_with_c}
\end{equation}
We further discuss hard constraints and provide concrete examples in \Cref{app:alg}.


\paragraph{Multi-timestep predictions.} We represent the programmatic experts and the world as distributions over the next-timestep observations in \cref{eq:rep}. It is likewise possible to reformulate a multi-timestep expert $p^{expert}(o_{t+1:t+H}|o_{1:t}, a_{1:t})$ as a product of next-timestep experts by assuming that the predictions of the multi-step expert at different timestep are independent:
\begin{equation}
    p^{expert}(o_{t+1:t+H}|o_{1:t}, a_{1:t}) = \prod_{k=1}^H p^{expert}(o_{t+k}|o_{1:t}, a_{1:t})
\end{equation}

\subsection{World Model Learning: Program Synthesis and Weight Optimization}\label{sec:learning}

We begin with a demonstration trajectory, learn a world model, and then begin to act in the world according to that model.
As the agent acts, it collects more trajectory data, which it uses to update or ``debug'' its model.
Concretely, learning proceeds as follows:
\begin{enumerate}[label=Step \arabic*:,leftmargin=1.8cm]
    \item Synthesize programmatic experts $\{p_i^{expert}\}_{i=1}^m$ given observed trajectories $D=\{\tau_i\}_{i=1}^n$
    \item Fit the weights $\btheta$ of the experts according to \cref{eq:mle} with a gradient-based optimizer
    \item Remove the experts with weights below threshold $\delta$
    \item Repeat Step 1-3 every time the observed trajectories get updated
\end{enumerate}

\paragraph{Generating program experts.} Following previous works \cite{tang2024worldcoder, dainese2024codeworldmodel, ahmed2025synthesizing}, we adopt Large Language Models (LLMs) as our Python program generator. We input a small batch of transitions $(o_{t:t+H+1}, a_{t:t+H})$ to the LLM prompt to produce the programmatic experts $\{p^{expert}_i\}$.

\Cref{fig:heatmap} shows how we interpret small Python programs as distributions over the observations. 
While we could ask LLMs to synthesize probabilistic programs to specify the distributions, we find it much more effective to have LLMs synthesize simple, deterministic Python programs, presumably because generic Python code is far more prevalent in LLM training data. 
We assume that an observation is represented as a list of objects, where each object has the following attributes: x/y velocity and visibility.
Then, a distribution over observations is a distribution over each object's attributes.
To interpret a Python program as a distribution, we assume that all object attributes are conditionally independent given full history, as mentioned in \Cref{sec:rep}, and convert all object attributes set by the program to distributions with single peaks at the given values. We then add noise to the distributions to ensure non-zero probabilities over alternative values. 
Any object attributes whose values are not set by the program follow uniform distributions over all possible values: consequently,
experts with a single if-condition (\cref{fig:heatmap}) yield a uniform distribution when the if-condition is not satisfied.

\paragraph{Gradient-based Weights Optimization.}

Once we have $\{p^{expert}_i\}_{i=1}^m$, we can perform maximum likelihood estimation to obtain $\btheta$:
\begin{equation}
    \btheta^* = \argmax_{\btheta} \sum_{ (o_{1:T+1}, a_{1:T})\in D} \sum_{t =1}^T \log p_\btheta (o_{t+1}|o_{1:t}, a_{1:t})
    \label{eq:mle}
\end{equation}
This equation instantiates \cref{eq:erm} by letting $p_{model}$ have a parametric form $p_{model} = p_\btheta$ and choosing negative log likelihood as the loss function. We can use any gradient-based optimizer to optimize the weights. We use L-BFGS \cite{liu1989limited}, which worked better than Adam \cite{kingma2014adam} and SGD, because we have small data and few parameters. 

Finally, the experts whose weights are below a threshold $\delta$ are removed from the world model. We repeat this loop every time there are new observations. 
\Cref{app:alg} contains full algorithm details.



\subsection{World Model Usage: RL in Simulation and Planning with World Model}



An important goal of world modeling is to aid decision-making. We consider two such ways of using world models. First, a world model can serve a simulator for reinforcement learning.
This makes RL policy learning more sample efficient, because we can quickly learn a world model from real environment interactions, which then substitutes the actual environment.
Subsequent policy learning can take place in the world model, obviating the need for further interaction with the real environment.
In practice, we also continue reinforcement learning after pretraining in the world model.
Formally, we learn a policy $\pi: \mathcal{O}^* \rightarrow \mathcal{A}$ that inputs an observation history and outputs an action.



\begin{figure*}[t!]
\centering
\includegraphics[width=0.99\linewidth]{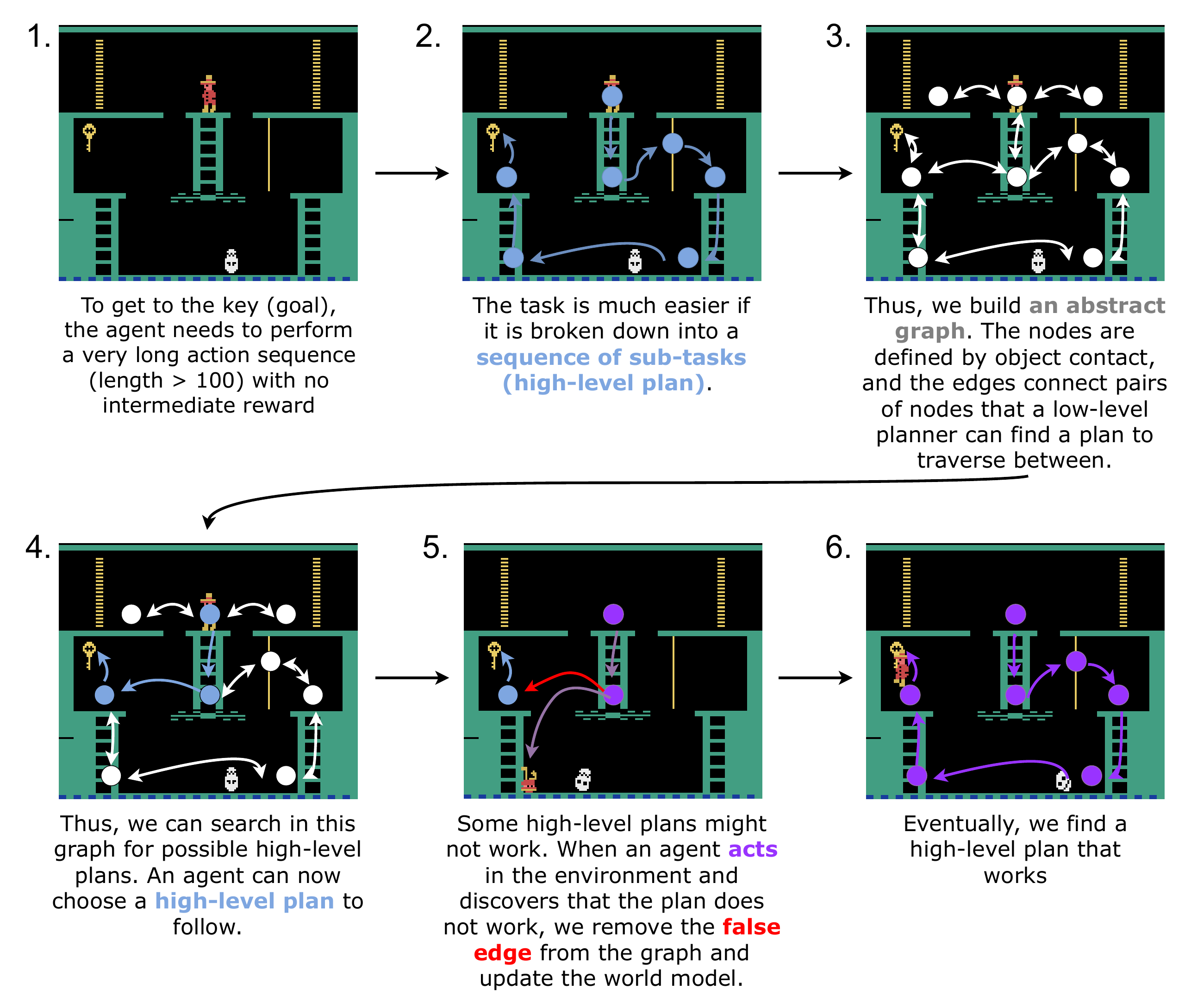}
\caption{A sequence of illustrations that demonstrates how our hierarchical planner works.}
\label{fig:planning}
\end{figure*}

Alternatively, world models can be used for lookahead planning. Given a reward function $R$, we plan for a horizon of $H$ timesteps by searching 
for an optimal action sequence, given our previous observations $o_{1:t}$ and actions $a_{1:t-1}$:
\begin{equation*}
    a_{t:t+H}^* = \argmax_{a_{t:t+H}} \mathbb{E}_{p_\btheta(o_{t+1:t+H}|o_{1:t}, a_{1:t+H})} \left[\sum_{k=0}^{H-1} R(o_{t+k+1}; o_{1:t+k}, a_{1:t+k})\right] 
\end{equation*}
where $p_\btheta(o_{t+1:t+H}|o_{1:t}, a_{1:t+H})=\prod_{k=0}^{H-1} p_\btheta(o_{t+k+1}|o_{1:t+k}, a_{1:t+k})$.

To help our world models 
 guide long-horizon decision-making, 
 we implement a hierarchical planner
inspired by task and motion planning (TAMP).
It first plans in a high-level abstract state space defined by object contact, then lowers those plans into actual Atari button presses using the learned world model (\Cref{fig:planning}, \Cref{app:planner}).
The resulting agent we denote \textbf{\method + Planner}.

\section{Experimental Results}

\begin{figure*}[b]
\centering
\vspace{-1em}
\includegraphics[width=0.85\linewidth]{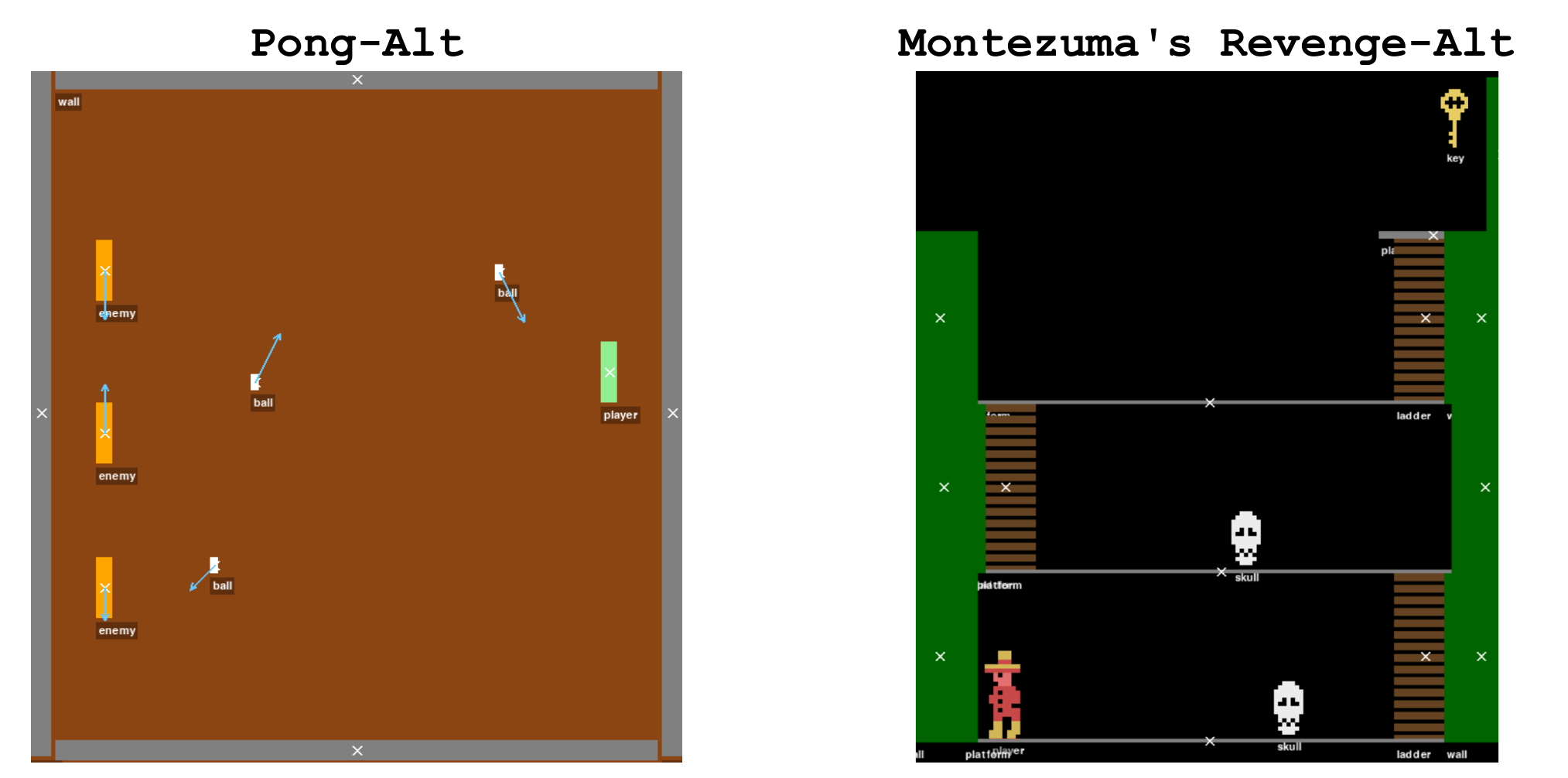}
\caption{Screenshots of the alternative environments. Pong-Alt's objective is to hit three balls past three enemies to score points, while Montezuma's Revenge-Alt requires players to avoid moving skulls by jumping over them and climbing up ladders to reach and collect the key.}
\label{fig:alt_env}
\end{figure*}

\paragraph{Domains and Evaluation.} We evaluate our agent, \textbf{\method + Planner}, against other methods on Atari's Pong and Montezuma's Revenge (MR), using the Arcade Learning Environment \cite{bellemare13atari}. 
We use OCAtari \cite{delfosse2023ocatari} to parse each image frame as a list of objects, each with an object category, a bounding box, and velocities. \footnote{OCAtari does not reliably extract objects so we have to make custom changes to OCAtari for every single Atari game we apply our method to, which makes it infeasible to run on the full Atari suite. We choose Pong and Montezuma's Revenge (arguably one of the hardest games in Atari---\cite{ecoffet2021first} calls it a ``grand challenge'') as representative games for which we manually patch OCAtari to reliably detect objects.} 
Both games are partially observed: the current state and action cannot uniquely determine the next state.
A demonstration of fewer than 1000 frames is created for each game, but 
these demonstrations are not successful gameplays: they serve only to illustrate the core causal mechanics.
In Montezuma's Revenge, our demonstration never achieves positive reward. 

To test compositional extrapolation
, we created 
alternative versions of both games, called Pong-Alt and Montezuma's Revenge-Alt (\Cref{fig:alt_env}), both of which recombine and rearrange the types of objects seen in the training demonstration.
Pong-Alt increases the number of objects (3 balls and 3 enemies). Montezuma's Revenge-Alt adds more enemies (which the player has to jump over) and ladders, while changing the map to resemble the game Kangaroo (see \Cref{app:domain} for details). We do not provide demonstrations for these alternative versions of the games. 

\paragraph{Baselines.} \textbf{PPO} \cite{schulman2017proximal} is a go-to, widely-used model-free RL algorithm. It optimizes a 
lower bound of the policy's performance using gradient descent. \textbf{LLM as Agent (ReAct)} \cite{yao2022react} directly uses an LLM as a policy. ReAct prompts LLMs to use extra chain-of-thought~\cite{wei2022cot} ``thinking'' actions before selecting an action.
\textbf{WorldCoder} \cite{tang2024worldcoder} is an LLM agent that models the world as a single Python program. It uses an LLM code generation and repair algorithm called REx \cite{tang2024code} to refine its world model to achieve high predictive accuracy on observed trajectories. More details in \Cref{app:baselines}.

\begin{figure*}[t]
\centering
\begin{tabular}{cc}
\includegraphics[width=0.46\linewidth]{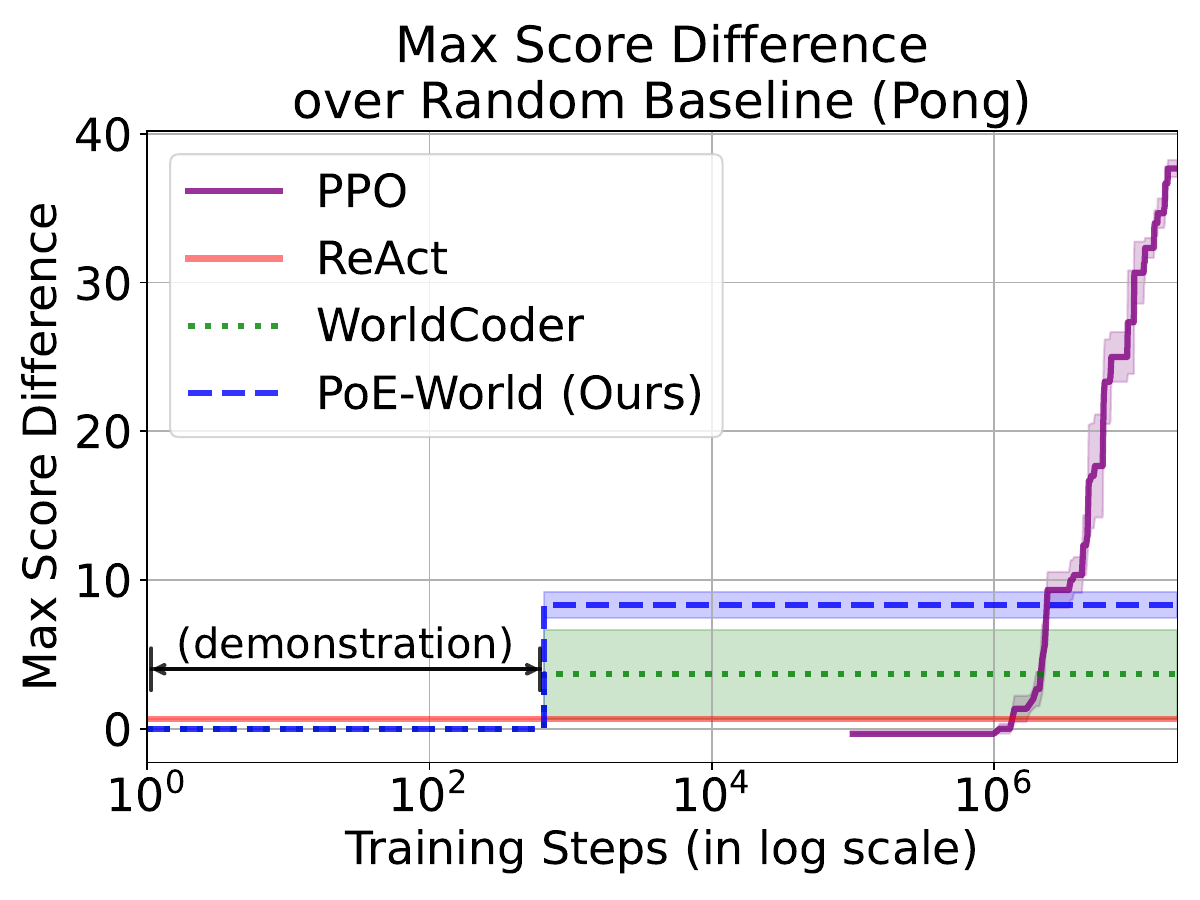} &
\includegraphics[width=0.46\linewidth]{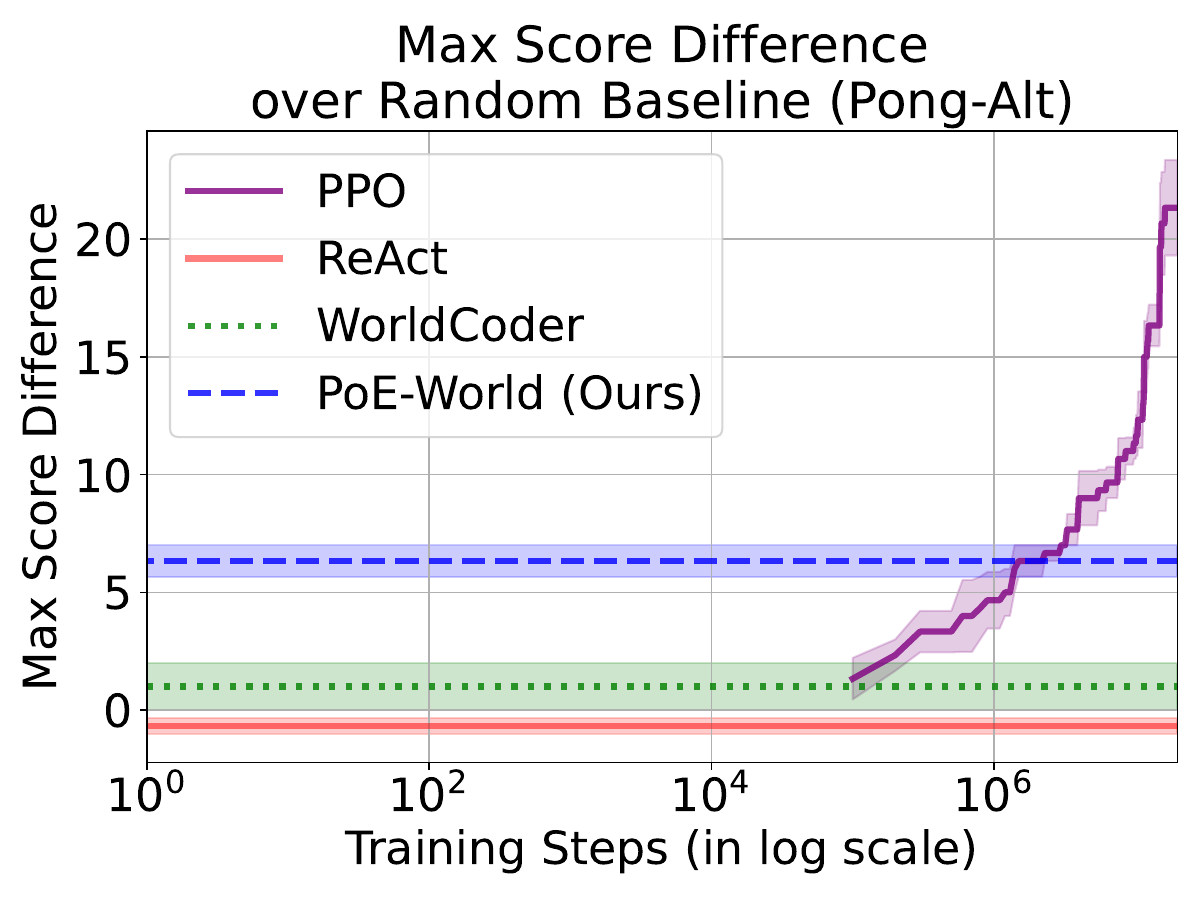}\\
\includegraphics[width=0.46\linewidth]{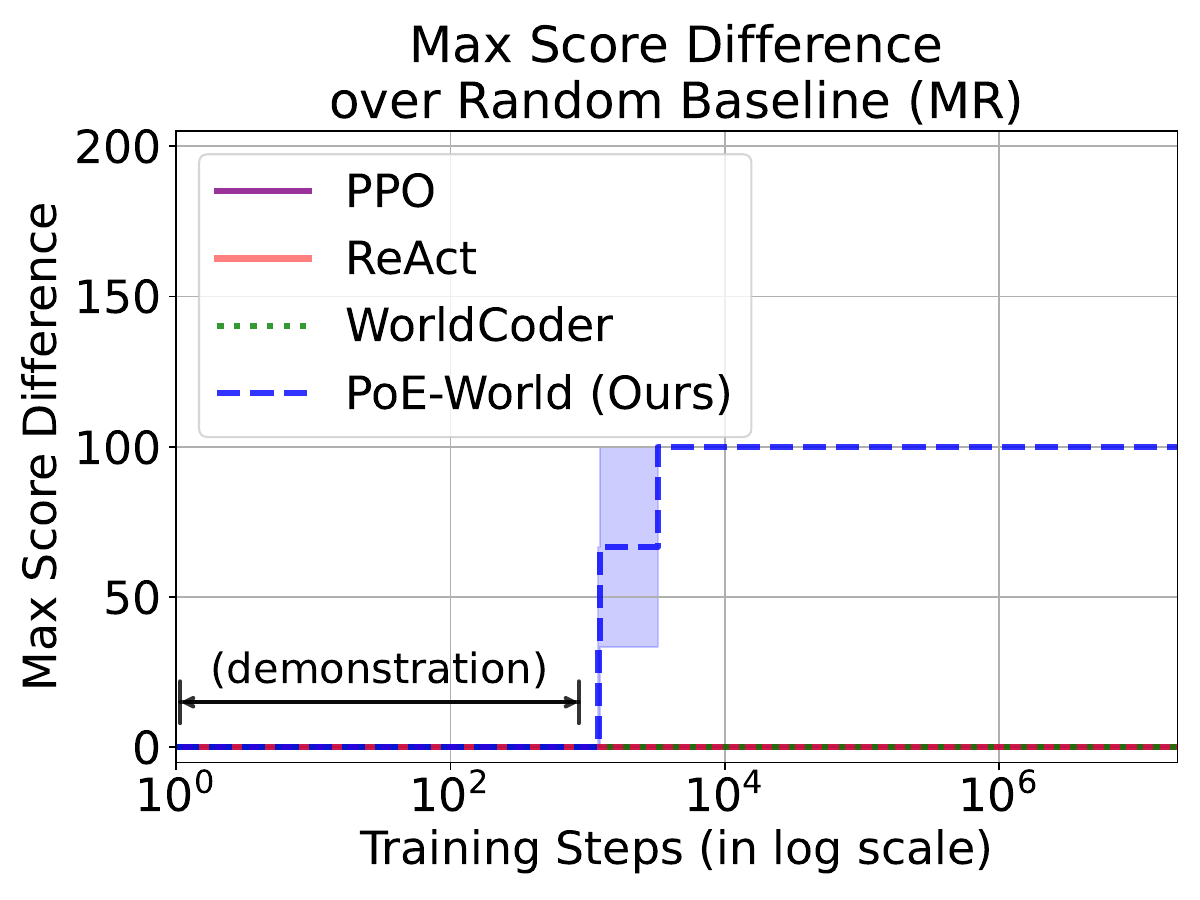} &
\includegraphics[width=0.46\linewidth]{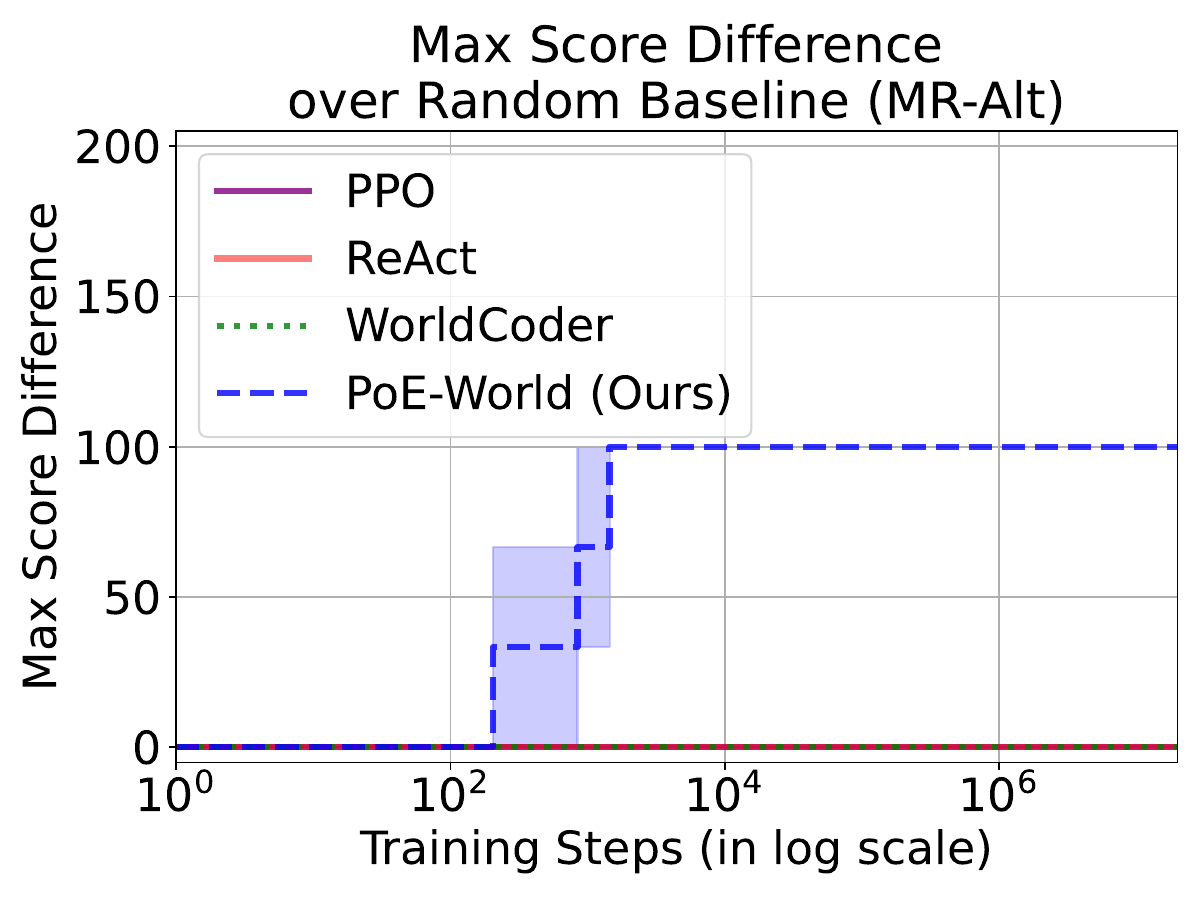} \\
\end{tabular}
\caption{Maximum score differences over random baseline achieved at different number of training steps (in log scale). The text "(demonstration)" annotates the length of demonstration used to initialize WorldCoder and \methodwospace, which is not displayed for the Alt games since we use the world models initialized by the demonstrations of the corresponding base games.}
\label{fig:score}
\end{figure*}
\begin{table*}[t]
\centering
\adjustbox{max width=\textwidth}{
\centering
\setlength\tabcolsep{4pt} 
\begin{tabular}{ccccccccc}
\toprule
Method & \multicolumn{4}{c}{Score}\\ 
\cmidrule(lr){2-5}
& Pong & Pong-Alt & MR & MR-Alt \\
\midrule
Random Agent & $-20.67 \pm 0.33$ & $-20.00 \pm 0.58$ & $0.00 \pm 0.00$ & $0.00 \pm 0.00$\\
PPO @ 100k env steps \cite{schulman2017proximal} & $-21.00 \pm 0.00$ & $-18.66 \pm 0.88$ & $0.00 \pm 0.00$ & $0.00 \pm 0.00$\\
LLM as Agent (ReAct) \cite{yao2022react} & $-20.00 \pm 0.00$ & $-20.67 \pm 0.33$ & $0.00 \pm 0.00$ & $0.00 \pm 0.00$\\
WorldCoder + Planner \cite{tang2024worldcoder} & $-17.00 \pm 3.00$ & $-19.00 \pm 1.00$ & $0.00 \pm 0.00$ & $0.00 \pm 0.00$\\
\method + Planner (Ours) & $\bm{-12.33 \pm 0.88}$ & $\bm{-13.67 \pm 0.67}$ & $\bm{100 \pm 0.00}$ & $\bm{100 \pm 0.00}$\\
\midrule
PPO @ 20m env steps \cite{schulman2017proximal} & $17.00 \pm 0.58$ & $1.33 \pm 2.03$ & $0.00 \pm 0.00$ & $0.00 \pm 0.00$\\
\bottomrule
\end{tabular}}
\caption{Scores on Pong and Montezuma's Revenge (MR) and their alternate versions. For PoE-World and WorldCoder, brief demonstrations on Pong and MR are given to initialize the world models. Their agents then train for at most 3k steps before evaluation.}
\vspace{0em}
\label{tab:score}
\end{table*}

\paragraph{Agent Results.} \Cref{fig:score} and \Cref{tab:score} show the scores of different agents on Pong, Pong-Alt, Montezuma's Revenge, and Montezuma's Revenge-Alt. The scores on Pong and Pong-Alt indicates the difference in points achieved by the player and the enemies when the game ends at 21 points. The scores on Montezuma's Revenge and Montezuma's Revenge-Alt become positive if and only if the agent succeeds in collecting the key.  As shown in \Cref{tab:score}, our agent, \textbf{\method + Planner}, performs best across all environments in the low-data regime, particularly when PPO baseline is allowed to have $100,000$ training environment interactions, the standard budget for sample-efficient agents on Atari 2600 \cite{kaiser2020atari100k}.
In \Cref{fig:score}, we keep training PPO for more steps, finding that it takes over a million steps for PPO to surpass \method + Planner. Moreover, PPO with 20M training steps never achieves positive score on Montezuma's Revenge. \method + Planner is the only method that manages to obtain positive reward on Montezuma's Revenge in both base and alternative versions. 

\pagebreak

\begin{wrapfigure}[15]{r}{0.5\linewidth} 
\vspace{-0.4em}
\centering
\includegraphics[width=\linewidth]{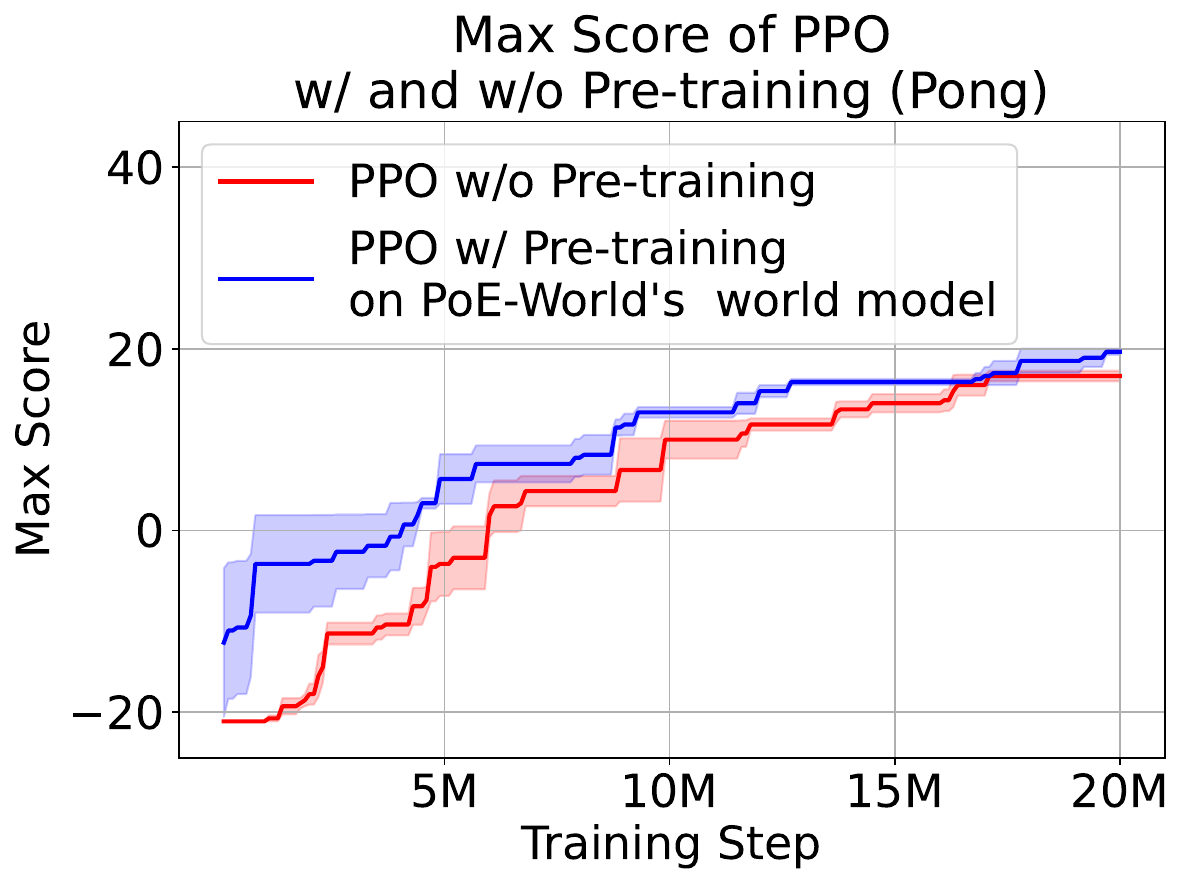}
\caption{
PPO with and without pre-training within our world model.}
\label{fig:pretrained}
\end{wrapfigure}

\paragraph{Instead of planning, can we use \method to learn a policy?} 
Training a policy avoids the test-time compute cost of planning.
In \Cref{fig:pretrained}, we show that pre-training a policy inside \methodwospace's world model  accelerates policy learning:
we first run PPO ``in simulation'' (in our world model), and then fine-tune in the actual Atari environment.
The fine-tuned PPO achieves significantly higher score than vanilla PPO at most training steps, and while the vanilla PPO takes 1M training steps to do better than a random agent, the fine-tuned PPO takes only 200k training steps. 
Asymptotically the pretrained and randomly-initialized policies converge to the same value:
World-model pretraining is effectively a way of warm-starting policy training.

\paragraph{Next Observation Prediction Results.} 
\Cref{tab:acc} and \Cref{tab:acc_partial} show next observation and next observation's object attributes prediction accuracies under the symbolic world modeling approaches, respectively. 
\method outperforms the baselines on most settings, except for the test observation (random frames) for Pong where all methods perform similarly because a random agent rarely succeeds in hitting the ball, and so knowledge of game mechanics is not comprehensively tested.

\begin{table*}[t]
\centering
\adjustbox{max width=\textwidth}{
\centering
\setlength\tabcolsep{4pt} 
\begin{tabular}{ccccccc}
\toprule
Method & \multicolumn{6}{c}{Next Observation Prediction Accuracy}\\
\cmidrule(lr){2-7}
& \multicolumn{2}{c}{Train} & \multicolumn{4}{c}{Test (1000 Random Frames)}\\
\cmidrule(lr){2-3} \cmidrule(lr){4-7}
& Pong & MR & Pong & Pong-Alt & MR & MR-Alt \\
\midrule
WorldCoder & $0.33 \pm 0.01$ & $0.36 \pm 0.02$ & $0.06 \pm 0.01$ & $0.01 \pm 0.01$ & $0.10 \pm 0.01$ & $0.12 \pm 0.03$\\
\method (ours) & $\bm{0.51 \pm 0.01}$ & $\bm{0.75 \pm 0.00}$ & $0.06 \pm 0.01$ & $\bm{0.03 \pm 0.00}$ & $\bm{0.31 \pm 0.00}$ & $\bm{0.43 \pm 0.00}$\\
\bottomrule
\end{tabular}}
\caption{Next observation prediction accuracy on the training demonstration and 1000 random frames.}
\label{tab:acc}
\end{table*}
\begin{table*}[t!]
\centering
\adjustbox{max width=\textwidth}{
\centering
\setlength\tabcolsep{4pt} 
\begin{tabular}{ccccccc}
\toprule
Method & \multicolumn{6}{c}{Next Observation's Object Attributes Prediction Accuracy}\\
\cmidrule(lr){2-7}
& \multicolumn{2}{c}{Train} & \multicolumn{4}{c}{Test (1000 Random Frames)}\\
\cmidrule(lr){2-3} \cmidrule(lr){4-7}
& Pong & MR & Pong & Pong-Alt & MR & MR-Alt \\
\midrule
WorldCoder & $0.82 \pm 0.00$ & $0.75 \pm 0.01$ & $0.66 \pm 0.01$ & $0.67 \pm 0.02$ & $0.58 \pm 0.00$ & $0.65 \pm 0.02$\\
\method (ours) & $\bm{0.88 \pm 0.00}$ & $\bm{0.93 \pm 0.00}$ & $0.66 \pm 0.00$ & $\bm{0.76 \pm 0.00}$ & $\bm{0.76 \pm 0.00}$ & $\bm{0.83 \pm 0.00}$\\
\bottomrule
\end{tabular}}
\caption{Next observation's object attributes prediction accuracies on the training demonstration and 1000 random frames. Non-moving objects, such as walls, platform, etc. are excluded.}
\label{tab:acc_partial}
\end{table*}
\begin{table*}[t!]
\centering
\adjustbox{max width=\textwidth}{
\centering
\setlength\tabcolsep{4pt} 
\begin{tabular}{ccccc}
\toprule
Method & \multicolumn{2}{c}{Planning successes} & \multicolumn{2}{c}{Next Observation Prediction Accuracy (Test)}\\ 
\cmidrule(lr){2-3} \cmidrule(lr){4-5}
& MR & MR-Alt & MR & MR-Alt \\
\midrule
\method + Planner w/o Hard Constraints & $2/9$ & $2/9$ & $0.30 \pm 0.01$ & $0.43 \pm 0.00$\\
\method + Planner & $\bm{5/9}$ & $\bm{4/9}$ & $0.31 \pm 0.00$ & $0.43 \pm 0.00$\\
\bottomrule
\end{tabular}}
\caption{Planning successes (out of 9 tries), where the goal is for the player to grab the key, and next observation prediction accuracies on 1000 random test frames of our agent with and without hard constraints on Montezuma's Revenge (MR) and its alternate version (MR-Alt).}
\label{tab:constraints}
\end{table*}

\paragraph{The role of hard constraints.} We investigate the role of hard constraints in our world model representation in \Cref{tab:constraints}.
In practice, hard constraints act to rule out ``physically impossible'' scenes. \method ended up enforcing constraints just for MR, since most possible scenes in Pong are already physically possible. 
As shown in \Cref{tab:constraints}, removing hard constraints 
causes the agent's performance to drop on both the base and alternative versions of Montezuma's Revenge. 
Interestingly, we found no significant changes in next observation prediction accuracies. We hypothesize that hard constraints do not necessarily turn bad predictions into good ones. Instead, they perform damage control---they refine poor predictions just enough to make them usable for long-horizon planning.

\paragraph{Qualitative difference between WorldCoder's and \methodwospace's world models.}  

World models produced by \method consists of 4000+ lines of code for Montezuma's Revenge compared to WorldCoder's less than 100 lines. 
This  difference is reflected in their ability to capture the underlying causal laws: \method models accurately represent important causal laws like character movement constraints with respect to platforms and ladders, while WorldCoder fails to model these mechanics, predicting that the player can ``fly'' around the map without any constraints. Moreover, WorldCoder models tend to hallucinate, e.g., imagining nonexistent bullet firing abilities in Montezuma's Revenge, potentially because there is no  granular way to downweigh buggy parts of a world model. \methodwospace, in contrast, prunes irrelevant experts 
with low weights, yielding more precise  world models.


\section{Related Work}

\textbf{World models as programs} has been explored in several recent works that motivate \methodwospace.
Similar to our work, WorldCoder~\cite{tang2024worldcoder} and CodeWorldModels~\cite{dainese2024codeworldmodel} use LLMs to write a Python transition function.
Unlike our work, they learn a single monolithic program, which severely limits scalability: our world models have an order of magnitude more code needed to model complex environments, as well as handle partial observability and non-determinism.
Recent works have also explored synthesizing high-level abstract world models as programs~\cite{liang2024visualpredicator,silver2021learning} to  support robotic planning by integrating visual perception with symbolic reasoning.
They could synergize with our work, as we have focused on learning low-level world models describing the motion of objects.

Earlier works~\cite{tsividis2021humanlevelreinforcementlearningtheorybased,10.1145/3571249} learn world models in restricted (non-Turing complete) domain specific programming languages, performing well on benchmarks co-designed with their Domain Specific Language.
Even earlier, Schema Networks learn a conceptually related factor-graph world model~\cite{kansky2017schema}.
AIXI~\cite{hutter2005universal} is a theoretical model of reinforcement learning which considers all possible Turing machines, mathematically related to all these works.
Interestingly, AIXI also works with a history space as a conceptually elegant way of modeling partial observability.
Other studies generate programs for world models primarily from natural language instructions, rather than from example interactions with the environment~\cite{sun2024factorsim,zhu2024bootstrapping,wong2023learning,guan2023leveraging}.
 
\textbf{Hierarchical Planning.} Our approach to segmenting complex continuous tasks into symbolic states is inspired by hierarchical~\cite{kaelbling2011hierarchical} and task-and motion~\cite{curtis2024partially,curtis2022discovering} planning.
Hierarchical representations in Reinforcement Learning (RL) are often expressed as options---temporal abstractions that allow agents to reason at multiple time scales by grouping sequences of primitive actions~\cite{sutton1999between}. 
Options can be manually specified by design or learned from experience \cite{bacon2017option}. 
Task and Motion Planning (TAMP) automatically discover symbolic states and action abstractions that enable generalizable symbolic plans grounded in continuous motion, however the applications of TAMP to Atari-style domains are still lacking due to complex dynamics.
Similar to recent work \cite{delfosse2023ocatari,bluml2025deep}, we define a high-level state space in terms of object contact relations---which is key to supporting symbolic abstraction.

\nocite{singh2012predictive}

\textbf{Alignment with human cognition.}
Our system, composed of simpler, specialized programmatic experts that give rise to complex behavior, is inspired by a view of mind as a community of interacting agents —- a recurring theme in philosophy and cognitive science~\cite{minsky1986society, dennett1993consciousness, lake2017building,marcus2003algebraic}.
Our modeling approach to modeling objects and actions aligns with empirical studies of event segmentation ~\cite{nguyen2024modeling,zacks2020event} and hierarchical planning~\cite{kryven2024approximate,balaguer2016neural}. Hierarchical state-spaces based on motion cues, such as ours, predict how people draw~\cite{tian2020learning} and interpret social interactions~\cite{shuadventures}, attesting to the cognitive alignment of our approach. Likewise, our programmatic representations of actions align with studies that demonstrate human concept learning to be akin to mental programs \cite{cogsci2025a,mapinference,lake2015human,lake2020people}, which recent work models by LLM-based program synthesis.


\section{Discussion and Limitations}

\paragraph{Compositionality.}
Recombining pieces of knowledge to generalize and extrapolate is a core feature of symbolic systems and, arguably, also of human cognition, ranging from natural language to abstract reasoning to everyday thought~\cite{dziri2023faith,lake2017building,lake2023human,spelke2007core}.
Our approach is in this spirit.
It factors its knowledge into small experts whose predictions can combine to extrapolate to scenes with more objects recomposed into new arrangements. 
More formally, our approach generalizes to novel "entity compositions" and "relational compositions," terms used by \cite{sehgal2023neurosymbolic}.
It should be noted, however, that our compositional factoring is orthogonal to our use of symbolic code:
Programs can be monolithic~\cite{tang2024worldcoder}, and neural nets can be factored~\cite{goyal2021recurrent}.
Nonetheless, compositionality proved critical to scalably learning the underlying symbolic program.
Synthesizing a single monolithic program is, in our view, intractable not just for the real world, but even for Atari.

\paragraph{Limitations.}
We make important assumptions, and only address part of the full model-based reinforcement learning problem.
Symbolic programs expect symbolic inputs:
We do not learn straight from pixels.
RL involves exploration, decision-making, and reward function learning, but we do not address those problems here.
However, we speculate our approach could unlock better methods for exploration:
A program-structured world model exposes an interpretable interface for describing beliefs about how the world works, and efficiently exploring the world is analogous to testing the program that encodes the world model.
Therefore ideas from software testing and program analysis could, in theory, be brought to bear, enabling new approaches to model-based exploration.


\nocite{machado18arcade}
\nocite{wang2023hypothesis}
\nocite{qiu2023phenomenal}
\nocite{piriyakulkij2024doing}
\nocite{acquaviva2024overcoming}
\nocite{bruce2024genie}
\nocite{hinton2002training}
\nocite{kansky2017schema}
\nocite{ecoffet2021first}
\nocite{garrett2021integrated}
\nocite{zhou2024robodreamer}
\nocite{du2020compositional}
\nocite{yang2023diffusion}
\nocite{ruoss2025lmact}

\begin{ack}
We thank Edward Gu and Taha Jafry for their contributions to earlier versions of this project. We are also grateful to Atharv Sonwane, Yilun Du, and Tom Silver for their valuable discussions and feedback on the paper, and we appreciate the OCAtari team's assistance and support with the OCAtari package. This work was supported by an NSF CAREER grant.
\end{ack}

\clearpage



\bibliographystyle{unsrt}
{\small \bibliography{CogSci_Template}}

\clearpage


\newpage
\appendix

\section{Appendix}

\subsection{\method algorithm details}\label{app:alg}

As mentioned in \Cref{sec:learning}, our learning algorithm \method alternates between two main steps: programmatic expert synthesis and gradient-based weight optimization. Given a trajectory, we do batch processing of size 10---we first learn $p_\theta(o_{11}|o_{1:10}, a_{1:10})$, then $p_\theta(o_{21}|o_{1:20}, a_{1:20})$, and so on. 

\begin{table*}[b]
\centering
\setlength\tabcolsep{4pt} 
\begin{tabular}{m{\linewidth}}
\toprule
\begin{spverbatim}
Example input list of objects:
player object (id = 0) with x-axis velocity = +4,
Interaction -- player object (id = 0) is touching ladder object (id = 2),
Interaction -- player object (id = 0) is touching unknown object (id = 4),

Example list of actions:
NOOP, NOOP, RIGHT

Example output list of object changes:
- The player object (id = 0) sets x-axis velocity to [+0, +0, +2]
\end{spverbatim}\\
\bottomrule
\end{tabular}
\caption{Example text representation of a sequence of consecutive transitions.}
\label{prompt:text_rep}
\end{table*}

\paragraph{Programmatic expert synthesis.} To synthesize programmatic experts, we first need to turn the observations into natural language observations that can be input into LLMs. A sequence of consecutive observation transitions $(o_{t:t+H+1}, a_{t:t+H})$ can be transformed into a text showing the first input object list, the list of actions, and the changes to the input object list for the following timesteps, all in natural language. \Cref{prompt:text_rep} shows an example text representation.

We note that in the text representation, only changes to a single object type (which is 'player' in \Cref{prompt:text_rep}) is displayed at a time. This choice is to enforce modularity between the object types: each experts put non-uniform distribution only on a single attribute of objects with a specific object type. This means \Cref{eq:rep} can be further rewritten as: 
\begin{equation}
    p_\btheta(o_{t+1}|o_{1:t}, a_{1:t}) \propto \prod_{obj-type}\prod_i p^{obj-type_\_expert}_i(o_{t+1}|o_{1:t}, a_{1:t})^{\theta_i}
    \label{eq:rep-appendix}
\end{equation}
where $\{p^{obj-type_\_expert}_i\}$ are the experts associated with each object type.

We then implement multiple (10 in total) LLM-based synthesis modules. For each object type, these modules take as input a sequence of consecutive observation transitions and output a set of programmatic experts. The output programmatic experts are pooled into a single set once all modules finish running. Having multiple synthesis modules allows each module to focus on different aspects of the environment, e.g., how objects move passively, how object moves when interacts with objects, how objects get created and deleted, etc. We describe one of our modules, ActionSynthesizer, below.

\begin{table*}[b]
\centering
\setlength\tabcolsep{4pt} 
\begin{tabular}{m{\linewidth}}
\toprule
\begin{spverbatim}
I'll give you an input list of objects and an output list of object changes, and I want you to list 4 possible reasons for the effects

Here's an example with player objects:
Example input list of objects:
player object (id = 0) with x-axis velocity = +0 and y-axis velocity +2,
Interaction -- player object (id = 0) is touching ladder object (id = 2),
Interaction -- player object (id = 0) is touching unknown object (id = 4),

Example output list of object changes:
- The player object (id = 0) sets x-axis velocity to +0
- The player object (id = 0) sets y-axis velocity to -4

Example reasons:
1. The player objects that touch an unknown object set their x-axis velocity to +0
2. The player objects that touch an unknown object set their y-axis velocity to -4
3. The player objects that touch an ladder object set their x-axis velocity to +0
4. The player objects that touch an ladder object set their y-axis velocity to -4

Please output a list of 4 reasons of the {obj_type} objects for the following input and output list of objects.

Input list of objects:
{input}

Output list of object changes:
{effects}

Please follow these rules for your output:
1. make sure each reason only talks about one object change
2. do not talk about IDs
\end{spverbatim}\\
\bottomrule
\end{tabular}
\caption{A prompt in ActionSynthesizer used to get LLM to provide causal explanations of the object changes.}
\label{prompt:interpret}
\end{table*}
\begin{table*}[b]
\centering
\setlength\tabcolsep{4pt} 
\begin{tabular}{m{\linewidth}}
\toprule
\small{
\begin{spverbatim}
We observe that the possible effects of {action} on {obj_type} objects include
{obs_lst_txt}

We want to synthesize python functions that implements these effects. The format of the functions should be

def alter_{obj_type}_objects(obj_list: ObjList, action: str) -> ObjList:
    {obj_type}_objs = obj_list.get_objs_by_obj_type('{obj_type}') # get all Obj of obj_type '{obj_type}'
    for {obj_type}_obj in {obj_type}_objs: # {obj_type}_obj is of type Obj
        # You can assume {obj_type}_obj.velocity_x and {obj_type}_obj.velocity_y are integers.
        pass
    return obj_list

And here are the docstrings for relevant classes:

class RandomValues:
    Use this class to express the possibility of random values. Example x = RandomValues([x + 2, x - 2])

    Attributes:
        values (list[ints]): list of possible values

    Methods:
        __init__(values):
            Initialize an instance

class Obj:
    Attributes:
        id (int): id of the object
        obj_type (string): type of the object
        velocity_x (int | RandomValues): x-axis velocity of the object
        velocity_y (int | RandomValues): y-axis velocity of the object
        deleted (int | RandomValues): whether this object gets deleted (1 if it does and 0 if it does not)

    Methods:
        touches(obj: Obj) -> bool:
            Returns whether this Obj is touching the input obj (True/False)

class ObjList:
    Attributes:
        objs (list of Obj)

    Methods:
        get_objs_by_obj_type(obj_type: str) -> list[Obj]:
            Returns list of objects with the input obj_type

        create_object(obj_type: str, x: int, y: int) -> ObjList:
            Returns a new instance of ObjList with the new object (with obj_type, x, y) added
\end{spverbatim}}\\
\bottomrule
\end{tabular}
\caption{First half of a prompt in ActionSynthesizer used to turn the natural language causal explanations into programs.}
\label{prompt:explain_1}
\end{table*}
\begin{table*}[b]
\centering
\setlength\tabcolsep{4pt} 
\begin{tabular}{m{\linewidth}}
\toprule
\begin{spverbatim}
Please output {n} different alter_{obj_type}_objects functions that explains each of the {n} possible effects of action '{action}' following these rules:
1. Each function should make changes to one attribute -- this could be the x-axis position, y-axis position, creation of object, or deletion of object.
2. Always use RandomValues to set attribute values. If there are conflicting changes to an attribute, instantiate RandomValues with a list of all possible values for that attribute.
3. Use Obj.touches to check for interactions.
4. Avoid setting each attribute value for each {obj_type} object more than once. For example, use 'break' inside a nested loop.
5. You can assume the velocities of input objects are integers.
6. Please use if-condition to indicate that the effects only happen because of action '{action}'
Format the output as a numbered list.
\end{spverbatim}\\
\bottomrule
\end{tabular}
\caption{Second half of a prompt in ActionSynthesizer used to turn the natural language causal explanations into programs.}
\label{prompt:explain_2}
\end{table*}

ActionSynthesizer focuses on synthesizing experts that explain how each action affects objects when the objects are interacting (touching) other objects (see example experts of this kind in \Cref{fig:heatmap}). 
It takes in just a single transition $(o_{t:t+1}, a_t)$. It turns this transition into a text representation as discussed above (\Cref{prompt:text_rep}). 
Then, it prompts a LLM to output causal explanations for the object changes (see the prompt \Cref{prompt:interpret}). 
An example causal explanation is ``the player objects that touch an ladder object set their y-axis velocity
to -4''. 
With a set of causal explanations, we prompt a LLM to turn each of them into a program (see the prompt \Cref{prompt:explain_1} and \Cref{prompt:explain_2}). 

Other modules follow this template of first prompting a LLM for natural language causal explanations and then prompting a LLM to turn the explanations into programs. We refer the reader to our code at \url{https://github.com/topwasu/poe-world} for the implementations of other LLM-based module. 

The programs synthesized by LLMs use our manually-written helper classes: \texttt{Obj}, \texttt{ObjList}, \texttt{RandomValues}, and \texttt{SeqValues}. Their docstrings are passed to the prompt. 
\texttt{Obj} provides a method \texttt{Obj.touches} which is the function we manually build in to help determine object contact. \texttt{Obj} contains object attributes: object category, x/y position, x/y velocities. 
It also contains properties including \texttt{center\_x}, \texttt{center\_y}, \texttt{left\_side}, \texttt{right\_side}, which are calculated based on the position and velocities, and their setter methods actually modify the velocities under the hood. 
\texttt{ObjList} is a class that represents an object list. It has a method \texttt{ObjList.get\_objs\_by\_obj\_type} and \texttt{ObjList.create\_object}. Lastly, \texttt{RandomValues} and \texttt{SeqValues} are the classes used to mark values set by a LLM. The purpose of these two classes is further discussed below. 

\begin{table*}[b]
\centering
\setlength\tabcolsep{4pt} 
\begin{tabular}{m{\linewidth}}
\toprule
\begin{spverbatim}
# Example program 1
def alter_player_objects(obj_list: ObjList, action: str, touch_side=3, touch_percent=0.6) -> ObjList:
    if action == 'NOOP':
        player_objs = obj_list.get_objs_by_obj_type('player')
        conveyer_belts = obj_list.get_objs_by_obj_type('conveyer_belt')
        for player_obj in player_objs:
            for conveyer_belt in conveyer_belts:
                if player_obj.touches(conveyer_belt, touch_side, touch_percent):
                    player_obj.velocity_x = RandomValues([-1])
                    break
    return obj_list

# Example program 2
def alter_player_objects(obj_list: ObjList, action: str, touch_side=3, touch_percent=1.0) -> ObjList:
    if action == 'FIRE':
        player_objs = obj_list.get_objs_by_obj_type('player')
        platform_objs = obj_list.get_objs_by_obj_type('platform')
        for player_obj in player_objs:
            for platform_obj in platform_objs:
                if player_obj.touches(platform_obj, touch_side, touch_percent):
                    player_obj.velocity_y = RandomValues([-6])
                    break
    return obj_list

# Example program 3
def alter_player_objects(obj_list: ObjList, action: str, touch_side=3, touch_percent=0.3) -> ObjList:
    if action == 'RIGHTFIRE':
        player_objs = obj_list.get_objs_by_obj_type('player')
        platform_objs = obj_list.get_objs_by_obj_type('platform')
        for player_obj in player_objs:
            for platform_obj in platform_objs:
                if player_obj.touches(platform_obj, touch_side, touch_percent):
                    player_obj.velocity_y = SeqValues([-6, -7, -4, 0, 2, 6, 9])
                    break
    return obj_list
\end{spverbatim}\\
\bottomrule
\end{tabular}
\caption{Example synthesized programmatic experts.}
\label{prompt:example_programs}
\end{table*}

After the programmatic experts are synthesized (see examples in \Cref{prompt:example_programs}), we interpret them as distributions as discussed in \Cref{sec:learning}. We tell LLMs in the prompt to set attribute values as instances of \texttt{RandomValues}, as opposed to integers. This marks the attributes whose values changed by a LLM so that we can write an program-to-distribution interpreter that puts a single-peak distribution on the attribute whose value is set and uniform distributions on all other attributes. \texttt{SeqValues} is similar to \texttt{RandomValues}, but is used in multi-timestep predictions scenario, as discussed in \Cref{sec:rep}.

The LLM used in the steps above is gpt-4o-2024-08-06. 
We implemented a disk cache for the LLM responses to avoid paying multiple times for the same prompts and seeds.

\begin{table*}[b]
\centering
\setlength\tabcolsep{4pt} 
\begin{tabular}{m{\linewidth}}
\toprule
\small{
\begin{spverbatim}
# Constraint 1: Player's body must align with the center of the rope
def c1(obj_list: ObjList, _, touch_side=2, touch_percent=0.3) -> tuple:
    touch_ids, satisfied_ids = [], []
    player_objs = obj_list.get_objs_by_obj_type('player')  # get all Obj of type 'player'
    rope_objs = obj_list.get_objs_by_obj_type('rope')  # get all Obj of type 'rope'
    for player_obj in player_objs:  # player_obj is of type Obj
        for rope_obj in rope_objs:  # rope_obj is of type Obj
            if player_obj.touches(rope_obj, touch_side, touch_percent):
                touch_ids.append(player_obj.id)
                if player_obj.center_x == rope_obj.center_x:
                    satisfied_ids.append(player_obj.id)
    return touch_ids, satisfied_ids

# Constraint 2: Player's feet must align with the top of the conveyer belt
def c2(obj_list: ObjList, _, touch_side=3, touch_percent=0.1) -> ObjList:
    touch_ids, satisfied_ids = [], []
    player_objs = obj_list.get_objs_by_obj_type('player')  # get all Obj of type 'player'
    conveyer_belt_objs = obj_list.get_objs_by_obj_type('conveyer_belt')  # get all Obj of type 'conveyer_belt'
    for player_obj in player_objs:  # player_obj is of type Obj
        for conveyer_belt_obj in conveyer_belt_objs:  # conveyer_belt_obj is of type Obj
            if player_obj.touches(conveyer_belt_obj, touch_side, touch_percent):
                touch_ids.append(conveyer_belt_obj.id)
                if player_obj.bottom_side == conveyer_belt_obj.top_side:
                    satisfied_ids.append(conveyer_belt_obj.id)
    return touch_ids, satisfied_ids

# Constraint 3: Player's feet must align with the top of the platform
def c3(obj_list: ObjList, _, touch_side=3, touch_percent=0.5) -> ObjList:
    touch_ids, satisfied_ids = [], []
    player_objs = obj_list.get_objs_by_obj_type('player')  # get all Obj of type 'player'
    platform_objs = obj_list.get_objs_by_obj_type('platform')  # get all Obj of type 'platform'
    for player_obj in player_objs:  # player_obj is of type Obj
        for platform_obj in platform_objs:  # platform_obj is of type Obj
            if player_obj.touches(platform_obj, touch_side, touch_percent):
                touch_ids.append(platform_obj.id)
                if player_obj.bottom_side == platform_obj.top_side:
                    satisfied_ids.append(platform_obj.id)
    return touch_ids, satisfied_ids

# Constraint 4: Player's body must align with the center of the ladder
def c4(obj_list: ObjList, _, touch_side=3, touch_percent=1.0) -> ObjList:
    touch_ids, satisfied_ids = [], []
    player_objs = obj_list.get_objs_by_obj_type('player')  # get all Obj of type 'player'
    ladder_objs = obj_list.get_objs_by_obj_type('ladder')  # get all Obj of type 'ladder'
    for player_obj in player_objs:  # player_obj is of type Obj
        for ladder_obj in ladder_objs:  # ladder_obj is of type Obj
            if player_obj.touches(ladder_obj, touch_side, touch_percent):
                touch_ids.append(ladder_obj.id)
                if player_obj.center_x == ladder_obj.center_x:
                    satisfied_ids.append(ladder_obj.id)
    return touch_ids, satisfied_ids
\end{spverbatim}}\\
\bottomrule
\end{tabular}
\caption{The collection of constraints synthesized for Montezuma's Revenge. Function names have been shortened to save space.}
\label{prompt:constraints}
\end{table*}

\paragraph{Hard constraints.} The hard constraints (\Cref{sec:rep}) are also synthesized similar to how we synthesize the programmatic experts. 
\Cref{prompt:constraints} shows the constraints learned for Montezuma's Revenge. 
In \Cref{eq:rep_with_c}, we choose to use a disjunction rather than a conjunction because the physics in video games can be peculiar and unrealistic---a player might have their body overlap with a platform when climbing down a ladder attached to that platform. In the real world, we believe a conjunction would be a better choice.

\paragraph{Gradient-based weight optimization.}

Once we have the expert distributions $\{p_i^{expert}\}$, we can optimize their weights $\{w_i\}$ according to \Cref{eq:mle}. 
We use the L-BFGS optimizer \cite{liu1989limited} implemented in PyTorch \cite{paszke2019pytorch} with strong Wolfe line search, learning rate = 1, number of epochs = 4, and without mini-batching. We also include a L1 regularization loss with weight = 1 so that the weights do not get too big.

We note that the weight optimization is done without taking into account the hard constraints since we would like to use gradient-based approaches.

We prune programs with weights lower than $\delta = 0.01$ after the weight optimization is done, and we prune constraints that contradict with the observations or explain less than 1\% of the observations.

\subsection{Planner}\label{app:planner}
 
In a game like Montezuma's Revenge, planning in the actual, low-level action space is hard because the number of actions required to get to the first positive reward is very high---at least 100. This means the size of search space is $8^{100}$ as there are 8 possible actions: NOOP, UP, DOWN, LEFT, RIGHT, FIRE, LEFTFIRE, RIGHTFIRE.

Thus, inspired by task and motion planning (TAMP), our hierarchical planner interleaves planning in the low-level motion action space with planning in a high-level abstract action space.
It learns an abstract graph where the nodes are abstract states defined by object contact, and the edges represent whether the world model believes the player can traverse between the two nodes.
Then, we search for a path in the abstract graph that takes the player to the goal.
In Atari, a goal is usually for a player to touch a goal object (key, ball, platforms, etc.).
The discovered path in the abstract graph is a high-level plan---a sequence of subgoals---for a low-level planning agent.
A low-level planning agent performs online planning to choose an action and then execute it in the environment. 

The planning algorithm can be described step-by-step as follows:
\begin{enumerate}[label=Step \arabic*:,leftmargin=1.8cm]
    \item Learn an abstract graph by running a low-level motion planner in simulation on all pairs of nodes. A (ordered) pair of nodes has an edge between them if we can find at least one low-level plan to traverse between them.
    \item Search for a path in the abstract graph. We use breadth-first search (BFS) here to find a path to the goal with the shortest length. The path is our high-level plan. If there is no path to the goal, go back to step 1.
    \item Attempt to follow the high-level plan, completing each subgoal in order, with a low-level planning agent.
    \item If there is a "false" edge, update the world model, remove that edge from the graph, then go back to step 2. Otherwise, the algorithm stops, and the agent has achieved the goal.
\end{enumerate}
\Cref{fig:planning} shows simplified illustrations of how our hierarchical planner works.

We now discuss the implementation of our low-level motion planner:

\paragraph{Low-level motion planner.} We implemented two low-level motion planners. The first is a variant of Monte Carlo Tree Search (MCTS) \cite{coulom2006efficient, browne2012survey}. It follows the same set of procedures as vanilla MCTS with two differences: first, similar to the MCTS algorithm used in \cite{tang2024worldcoder}, it approximates the value of a node using a heuristic function instead of doing a random rollout in the simulate step. The heuristic function is the Manhattan distance between the current position of the player object and the position of a goal object. We find that this function is a good estimate of how good an observation is when trying to achieve a goal. Second, the value of a node is updated as the maximum value of its children nodes, instead of the expected value, in the backpropagation step. Intuitively, using the maximum value encourages the planner to be more optimistic. The exploration parameter for MCTS is initially equal to 1 for Montezuma's Revenge and Montezuma's Revenge-Alt and 10 for Pong and Pong-Alt, and it increases by 10 times every 1000 iterations of MCTS.

We also implement ``sticky actions'' by extending the action space that MCTS searches on so that it includes repeated sequences of primitive actions with lengths $1$, $4$, and $8$. Thus, the extended action space has $3n$ actions where $n$ is the number of primitive actions. We include these repeated action chunks in our action space to make planning easier. Playing Atari games rarely requires players to change actions at every timestep, so repeated action chunks can be helpful.

The second low-level motion planner is a greedy search with the same heuristic function used in MCTS. At each iteration, it greedily finds the best repeated action chunk of length 8 and includes it in the plan.
It backtracks if the current state leads to death no matter which action chunk of length 8 the planner chooses.
It returns a plan when the player achieves the goal (touches the goal object).

Pong, Pong-Alt, and Montezuma's Revenge-Alt agents only use the greedy planner, while Montezuma's Revenge uses both: it first tries to find a plan with MCTS for 4000 iterations and falls back to the greedy planner if MCTS fails.

As discussed earlier, the low-level motion planner is used in two steps: to build the abstract graph by finding a plan to traverse between two abstract nodes in the world model, and to help inform a low-level planning agent that is trying to complete a subgoal.

\paragraph{Low-level planning agent.} The agent performs online planning: it uses the low-level planner to find a plan to the goal, and then it takes a sequence of actions and replans.
The agent replans when the ccurent plan no longer takes the agent to the goal, so the agent may take several actions before replanning.
We further optimize the agent by letting it replan only 40\% of the times when the current plan no longer works in Montezuma's Revenge.

Because of this introduced stochasticity, however, we find that our whole planning pipeline can give different scores in different runs even with the same initial world model, so we treat running the hierarchical planner multiple times with the same initial world model as part of training, and we run the planning algorithm 3 times on the same initial world model to get results on Montezuma's Revenge and Montezuma's Revenge-Alt. 
Our work focuses on world modeling, and we leave it to future work to increase the efficiency and performance of the hierarchical planner.

\subsection{Domain details}\label{app:domain}

\begin{figure*}[b]
\centering
\vspace{-1em}
\includegraphics[width=0.85\linewidth]{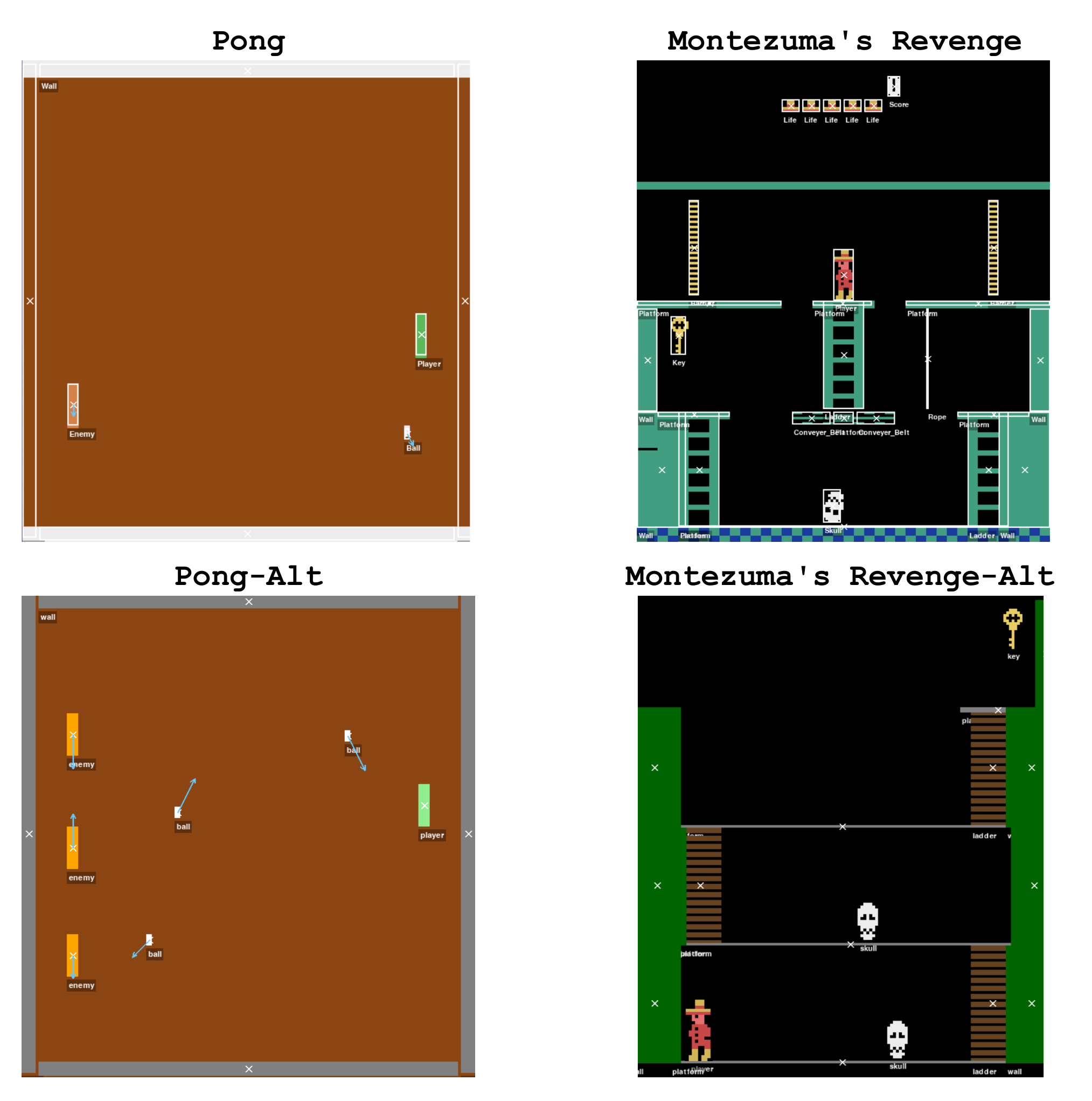}
\caption{Screenshots of both base and alternative environments for Pong and Montezuma's Revenge.}
\label{fig:all_env}
\end{figure*}

We evaluate our methods on Atari's Pong and Montezuma's Revenge using the Arcade Learning Environment (ALE). The frameskip parameter is set to 3 for both games. We use OCAtari to parse each image frame as a list of objects, each with
an object category, a bounding box, and velocities. 
OCAtari reverses-engineers the RAM values of Atari games to get the bounding boxes of each object. 

We modify OCAtari to fix a number of issues in its handling of Pong and Montezuma's Revenge. This includes fixing bugs in object detection so that it fully detects every object in play, correcting bugs in the the bounding boxes so that object interactions correctly correspond to when object bounding boxes touch/overlap,  etc. We refer the reader to our code implementation \url{https://github.com/topwasu/poe-world} for full details. 

Details on how we create Pong-Alt and Montezuma's Revenge-Alt are below:

\paragraph{Pong-Alt} is created by layering three Pong environments. We sync the player's location in all three environments, so that it appears as if we have only one paddle. The enemies and balls, on the other hand, are all at different locations. We end up with one player, three enemies, and three balls. 

\paragraph{Montezuma's Revenge-Alt} is created by stacking the lower section of the first room in the original Montezuma's Revenge to make three platforms of different heights, connected by stairs. We stack up three lower sections of three different Montezuma's Revenge environments. The first and the third section in the stack are flipped horizontally so that the stairs are on different sides of the room, requiring the player to jump over the skulls to reach the stairs. 

ALE code uses GPL-2.0 license, and OCAtari code uses MIT license.

\subsection{Baseline details}\label{app:baselines}

\paragraph{PPO.} PPO uses frame stacking = 4. We use the same hyperparameters as the PPO paper \cite{schulman2017proximal} and OCAtari \cite{delfosse2023ocatari}. We use the stable-baselines3 implementation of PPO with MLP backbone ('MlpPolicy') \cite{raffin2021stable}. 
For the Alt environments, we take the PPO model pretrained 20M steps on the base environments and finetune it on the Alt environments. 
This pretraining process is done for fiar comparison with our method, since our method assumes demonstrations from the base environments when evaluating on the corresponding Alt environments.

\paragraph{ReAct.} We write prompts that would alternate between thinking and taking actions, one for Pong and Pong-Alt and another for Montezuma's Revenge and Montezuma's Revenge-Alt. The frame observation is transformed into text where we provide each object's x and y position and a list of all pairwise object interactions. This text observation is input as part of the prompt. In the prompt, we only include the 4 most recent observations (along with the 4 most recent thinking actions and 4 most recent taken actions) as each observation is quite long in text.

\paragraph{WorldCoder.} We use the official WorldCoder implementation \cite{tang2024worldcoder} but replace the existing prompts with new ones that are tailored towards our text representation of object-centric Atari frames. The prompts can be found in our codebase \url{https://github.com/topwasu/poe-world}. The instantiation of WorldCoder in \cite{tang2024worldcoder} actually has its own planner implemented, but for fair comparison with our method, we use only the world modeling part of that system and combine it with our own planner so that the planner is the same for both WorldCoder and our method. We choose to use our own planner instead of theirs since ours is hierarchical. 

WorldCoder code uses MIT license.

\subsection{Compute resources and execution time}\label{app:cost}

\paragraph{Compute Resources.}

For the world modeling part, our experiments are run on 4 CPUs (CascadeLake, IceLake, or SaphireRapids) with 64 GB memory. \method and WorldCoder uses a budget of \$20 worth of OpenAI credit per run. For the planner, we also run it mostly on 4 CPUs, but for the part where we need to build an abstract graph by running many low-level planners, we parallelize it on multuple compute jobs on a job scheduling cluster, so we might be using 100 CPUs at a time.

\paragraph{Execution time.}

\method alone without the planner tends to take around 8 hours to run (this includes the time we need to wait for OpenAI LLM requests). The planner running time varies, but most runs finish under 24 hours.

\clearpage


\newpage
\section*{NeurIPS Paper Checklist}

\begin{enumerate}

\item {\bf Claims}
    \item[] Question: Do the main claims made in the abstract and introduction accurately reflect the paper's contributions and scope?
    \item[] Answer: \answerYes{} 
    \item[] Justification: The claims made in the abstract and introduction accurately reflect the paper's contributions and scope.
    \item[] Guidelines:
    \begin{itemize}
        \item The answer NA means that the abstract and introduction do not include the claims made in the paper.
        \item The abstract and/or introduction should clearly state the claims made, including the contributions made in the paper and important assumptions and limitations. A No or NA answer to this question will not be perceived well by the reviewers. 
        \item The claims made should match theoretical and experimental results, and reflect how much the results can be expected to generalize to other settings. 
        \item It is fine to include aspirational goals as motivation as long as it is clear that these goals are not attained by the paper. 
    \end{itemize}

\item {\bf Limitations}
    \item[] Question: Does the paper discuss the limitations of the work performed by the authors?
    \item[] Answer: \answerYes{} 
    \item[] Justification: The paper discusses the limitations of the work performed by the authors.
    \item[] Guidelines:
    \begin{itemize}
        \item The answer NA means that the paper has no limitation while the answer No means that the paper has limitations, but those are not discussed in the paper. 
        \item The authors are encouraged to create a separate "Limitations" section in their paper.
        \item The paper should point out any strong assumptions and how robust the results are to violations of these assumptions (e.g., independence assumptions, noiseless settings, model well-specification, asymptotic approximations only holding locally). The authors should reflect on how these assumptions might be violated in practice and what the implications would be.
        \item The authors should reflect on the scope of the claims made, e.g., if the approach was only tested on a few datasets or with a few runs. In general, empirical results often depend on implicit assumptions, which should be articulated.
        \item The authors should reflect on the factors that influence the performance of the approach. For example, a facial recognition algorithm may perform poorly when image resolution is low or images are taken in low lighting. Or a speech-to-text system might not be used reliably to provide closed captions for online lectures because it fails to handle technical jargon.
        \item The authors should discuss the computational efficiency of the proposed algorithms and how they scale with dataset size.
        \item If applicable, the authors should discuss possible limitations of their approach to address problems of privacy and fairness.
        \item While the authors might fear that complete honesty about limitations might be used by reviewers as grounds for rejection, a worse outcome might be that reviewers discover limitations that aren't acknowledged in the paper. The authors should use their best judgment and recognize that individual actions in favor of transparency play an important role in developing norms that preserve the integrity of the community. Reviewers will be specifically instructed to not penalize honesty concerning limitations.
    \end{itemize}

\item {\bf Theory assumptions and proofs}
    \item[] Question: For each theoretical result, does the paper provide the full set of assumptions and a complete (and correct) proof?
    \item[] Answer: \answerNA{} 
    \item[] Justification: The paper does not include theoretical results.
    \item[] Guidelines:
    \begin{itemize}
        \item The answer NA means that the paper does not include theoretical results. 
        \item All the theorems, formulas, and proofs in the paper should be numbered and cross-referenced.
        \item All assumptions should be clearly stated or referenced in the statement of any theorems.
        \item The proofs can either appear in the main paper or the supplemental material, but if they appear in the supplemental material, the authors are encouraged to provide a short proof sketch to provide intuition. 
        \item Inversely, any informal proof provided in the core of the paper should be complemented by formal proofs provided in appendix or supplemental material.
        \item Theorems and Lemmas that the proof relies upon should be properly referenced. 
    \end{itemize}

    \item {\bf Experimental result reproducibility}
    \item[] Question: Does the paper fully disclose all the information needed to reproduce the main experimental results of the paper to the extent that it affects the main claims and/or conclusions of the paper (regardless of whether the code and data are provided or not)?
    \item[] Answer: \answerYes{} 
    \item[] Justification: We have fully described the algorithms used in the paper, with details included in the appendix.
    \item[] Guidelines:
    \begin{itemize}
        \item The answer NA means that the paper does not include experiments.
        \item If the paper includes experiments, a No answer to this question will not be perceived well by the reviewers: Making the paper reproducible is important, regardless of whether the code and data are provided or not.
        \item If the contribution is a dataset and/or model, the authors should describe the steps taken to make their results reproducible or verifiable. 
        \item Depending on the contribution, reproducibility can be accomplished in various ways. For example, if the contribution is a novel architecture, describing the architecture fully might suffice, or if the contribution is a specific model and empirical evaluation, it may be necessary to either make it possible for others to replicate the model with the same dataset, or provide access to the model. In general. releasing code and data is often one good way to accomplish this, but reproducibility can also be provided via detailed instructions for how to replicate the results, access to a hosted model (e.g., in the case of a large language model), releasing of a model checkpoint, or other means that are appropriate to the research performed.
        \item While NeurIPS does not require releasing code, the conference does require all submissions to provide some reasonable avenue for reproducibility, which may depend on the nature of the contribution. For example
        \begin{enumerate}
            \item If the contribution is primarily a new algorithm, the paper should make it clear how to reproduce that algorithm.
            \item If the contribution is primarily a new model architecture, the paper should describe the architecture clearly and fully.
            \item If the contribution is a new model (e.g., a large language model), then there should either be a way to access this model for reproducing the results or a way to reproduce the model (e.g., with an open-source dataset or instructions for how to construct the dataset).
            \item We recognize that reproducibility may be tricky in some cases, in which case authors are welcome to describe the particular way they provide for reproducibility. In the case of closed-source models, it may be that access to the model is limited in some way (e.g., to registered users), but it should be possible for other researchers to have some path to reproducing or verifying the results.
        \end{enumerate}
    \end{itemize}

\item {\bf Open access to data and code}
    \item[] Question: Does the paper provide open access to the data and code, with sufficient instructions to faithfully reproduce the main experimental results, as described in supplemental material?
    \item[] Answer: \answerYes{} 
    \item[] Justification: The code is available at \url{https://anonymous.4open.science/r/poe-world-neurips-628F/}
    \item[] Guidelines:
    \begin{itemize}
        \item The answer NA means that paper does not include experiments requiring code.
        \item Please see the NeurIPS code and data submission guidelines (\url{https://nips.cc/public/guides/CodeSubmissionPolicy}) for more details.
        \item While we encourage the release of code and data, we understand that this might not be possible, so “No” is an acceptable answer. Papers cannot be rejected simply for not including code, unless this is central to the contribution (e.g., for a new open-source benchmark).
        \item The instructions should contain the exact command and environment needed to run to reproduce the results. See the NeurIPS code and data submission guidelines (\url{https://nips.cc/public/guides/CodeSubmissionPolicy}) for more details.
        \item The authors should provide instructions on data access and preparation, including how to access the raw data, preprocessed data, intermediate data, and generated data, etc.
        \item The authors should provide scripts to reproduce all experimental results for the new proposed method and baselines. If only a subset of experiments are reproducible, they should state which ones are omitted from the script and why.
        \item At submission time, to preserve anonymity, the authors should release anonymized versions (if applicable).
        \item Providing as much information as possible in supplemental material (appended to the paper) is recommended, but including URLs to data and code is permitted.
    \end{itemize}

\item {\bf Experimental setting/details}
    \item[] Question: Does the paper specify all the training and test details (e.g., data splits, hyperparameters, how they were chosen, type of optimizer, etc.) necessary to understand the results?
    \item[] Answer: \answerYes{} 
    \item[] Justification: Algorithm details are described in the appendix.
    \item[] Guidelines:
    \begin{itemize}
        \item The answer NA means that the paper does not include experiments.
        \item The experimental setting should be presented in the core of the paper to a level of detail that is necessary to appreciate the results and make sense of them.
        \item The full details can be provided either with the code, in appendix, or as supplemental material.
    \end{itemize}

\item {\bf Experiment statistical significance}
    \item[] Question: Does the paper report error bars suitably and correctly defined or other appropriate information about the statistical significance of the experiments?
    \item[] Answer: \answerYes{} 
    \item[] Justification: The paper reports error bars computed over 5 seeds.
    \item[] Guidelines:
    \begin{itemize}
        \item The answer NA means that the paper does not include experiments.
        \item The authors should answer "Yes" if the results are accompanied by error bars, confidence intervals, or statistical significance tests, at least for the experiments that support the main claims of the paper.
        \item The factors of variability that the error bars are capturing should be clearly stated (for example, train/test split, initialization, random drawing of some parameter, or overall run with given experimental conditions).
        \item The method for calculating the error bars should be explained (closed form formula, call to a library function, bootstrap, etc.)
        \item The assumptions made should be given (e.g., Normally distributed errors).
        \item It should be clear whether the error bar is the standard deviation or the standard error of the mean.
        \item It is OK to report 1-sigma error bars, but one should state it. The authors should preferably report a 2-sigma error bar than state that they have a 96\% CI, if the hypothesis of Normality of errors is not verified.
        \item For asymmetric distributions, the authors should be careful not to show in tables or figures symmetric error bars that would yield results that are out of range (e.g. negative error rates).
        \item If error bars are reported in tables or plots, The authors should explain in the text how they were calculated and reference the corresponding figures or tables in the text.
    \end{itemize}

\item {\bf Experiments compute resources}
    \item[] Question: For each experiment, does the paper provide sufficient information on the computer resources (type of compute workers, memory, time of execution) needed to reproduce the experiments?
    \item[] Answer: \answerYes{} 
    \item[] Justification: We have provided OpenAI API cost for our experiments and compute resources used for our experiments.
    \item[] Guidelines:
    \begin{itemize}
        \item The answer NA means that the paper does not include experiments.
        \item The paper should indicate the type of compute workers CPU or GPU, internal cluster, or cloud provider, including relevant memory and storage.
        \item The paper should provide the amount of compute required for each of the individual experimental runs as well as estimate the total compute. 
        \item The paper should disclose whether the full research project required more compute than the experiments reported in the paper (e.g., preliminary or failed experiments that didn't make it into the paper). 
    \end{itemize}
    
\item {\bf Code of ethics}
    \item[] Question: Does the research conducted in the paper conform, in every respect, with the NeurIPS Code of Ethics \url{https://neurips.cc/public/EthicsGuidelines}?
    \item[] Answer: \answerYes{} 
    \item[] Justification:  The research conducted in the paper conforms, in every aspect, with the NeurIPS Code of Ethics.
    \item[] Guidelines:
    \begin{itemize}
        \item The answer NA means that the authors have not reviewed the NeurIPS Code of Ethics.
        \item If the authors answer No, they should explain the special circumstances that require a deviation from the Code of Ethics.
        \item The authors should make sure to preserve anonymity (e.g., if there is a special consideration due to laws or regulations in their jurisdiction).
    \end{itemize}

\item {\bf Broader impacts}
    \item[] Question: Does the paper discuss both potential positive societal impacts and negative societal impacts of the work performed?
    \item[] Answer: \answerNA{} 
    \item[] Justification: Our work is foundational research and has no direct societal impact.
    \item[] Guidelines:
    \begin{itemize}
        \item The answer NA means that there is no societal impact of the work performed.
        \item If the authors answer NA or No, they should explain why their work has no societal impact or why the paper does not address societal impact.
        \item Examples of negative societal impacts include potential malicious or unintended uses (e.g., disinformation, generating fake profiles, surveillance), fairness considerations (e.g., deployment of technologies that could make decisions that unfairly impact specific groups), privacy considerations, and security considerations.
        \item The conference expects that many papers will be foundational research and not tied to particular applications, let alone deployments. However, if there is a direct path to any negative applications, the authors should point it out. For example, it is legitimate to point out that an improvement in the quality of generative models could be used to generate deepfakes for disinformation. On the other hand, it is not needed to point out that a generic algorithm for optimizing neural networks could enable people to train models that generate Deepfakes faster.
        \item The authors should consider possible harms that could arise when the technology is being used as intended and functioning correctly, harms that could arise when the technology is being used as intended but gives incorrect results, and harms following from (intentional or unintentional) misuse of the technology.
        \item If there are negative societal impacts, the authors could also discuss possible mitigation strategies (e.g., gated release of models, providing defenses in addition to attacks, mechanisms for monitoring misuse, mechanisms to monitor how a system learns from feedback over time, improving the efficiency and accessibility of ML).
    \end{itemize}
    
\item {\bf Safeguards}
    \item[] Question: Does the paper describe safeguards that have been put in place for responsible release of data or models that have a high risk for misuse (e.g., pretrained language models, image generators, or scraped datasets)?
    \item[] Answer: \answerNA{} 
    \item[] Justification: The paper poses no such risks.
    \item[] Guidelines:
    \begin{itemize}
        \item The answer NA means that the paper poses no such risks.
        \item Released models that have a high risk for misuse or dual-use should be released with necessary safeguards to allow for controlled use of the model, for example by requiring that users adhere to usage guidelines or restrictions to access the model or implementing safety filters. 
        \item Datasets that have been scraped from the Internet could pose safety risks. The authors should describe how they avoided releasing unsafe images.
        \item We recognize that providing effective safeguards is challenging, and many papers do not require this, but we encourage authors to take this into account and make a best faith effort.
    \end{itemize}

\item {\bf Licenses for existing assets}
    \item[] Question: Are the creators or original owners of assets (e.g., code, data, models), used in the paper, properly credited and are the license and terms of use explicitly mentioned and properly respected?
    \item[] Answer: \answerYes{} 
    \item[] Justification: The creators of original data are all properly credited, and the license and terms of use of the data are explicitly mentioned and properly respected.
    \item[] Guidelines:
    \begin{itemize}
        \item The answer NA means that the paper does not use existing assets.
        \item The authors should cite the original paper that produced the code package or dataset.
        \item The authors should state which version of the asset is used and, if possible, include a URL.
        \item The name of the license (e.g., CC-BY 4.0) should be included for each asset.
        \item For scraped data from a particular source (e.g., website), the copyright and terms of service of that source should be provided.
        \item If assets are released, the license, copyright information, and terms of use in the package should be provided. For popular datasets, \url{paperswithcode.com/datasets} has curated licenses for some datasets. Their licensing guide can help determine the license of a dataset.
        \item For existing datasets that are re-packaged, both the original license and the license of the derived asset (if it has changed) should be provided.
        \item If this information is not available online, the authors are encouraged to reach out to the asset's creators.
    \end{itemize}

\item {\bf New assets}
    \item[] Question: Are new assets introduced in the paper well documented and is the documentation provided alongside the assets?
    \item[] Answer: \answerNA{} 
    \item[] Justification: The paper currently does not release new assets.
    \item[] Guidelines:
    \begin{itemize}
        \item The answer NA means that the paper does not release new assets.
        \item Researchers should communicate the details of the dataset/code/model as part of their submissions via structured templates. This includes details about training, license, limitations, etc. 
        \item The paper should discuss whether and how consent was obtained from people whose asset is used.
        \item At submission time, remember to anonymize your assets (if applicable). You can either create an anonymized URL or include an anonymized zip file.
    \end{itemize}

\item {\bf Crowdsourcing and research with human subjects}
    \item[] Question: For crowdsourcing experiments and research with human subjects, does the paper include the full text of instructions given to participants and screenshots, if applicable, as well as details about compensation (if any)? 
    \item[] Answer: \answerNA{} 
    \item[] Justification: The paper does not involve crowdsourcing nor research with human subjects.
    \item[] Guidelines:
    \begin{itemize}
        \item The answer NA means that the paper does not involve crowdsourcing nor research with human subjects.
        \item Including this information in the supplemental material is fine, but if the main contribution of the paper involves human subjects, then as much detail as possible should be included in the main paper. 
        \item According to the NeurIPS Code of Ethics, workers involved in data collection, curation, or other labor should be paid at least the minimum wage in the country of the data collector. 
    \end{itemize}

\item {\bf Institutional review board (IRB) approvals or equivalent for research with human subjects}
    \item[] Question: Does the paper describe potential risks incurred by study participants, whether such risks were disclosed to the subjects, and whether Institutional Review Board (IRB) approvals (or an equivalent approval/review based on the requirements of your country or institution) were obtained?
    \item[] Answer: \answerNA{} 
    \item[] Justification: The paper does not involve crowdsourcing nor research with human subjects.
    \item[] Guidelines:
    \begin{itemize}
        \item The answer NA means that the paper does not involve crowdsourcing nor research with human subjects.
        \item Depending on the country in which research is conducted, IRB approval (or equivalent) may be required for any human subjects research. If you obtained IRB approval, you should clearly state this in the paper. 
        \item We recognize that the procedures for this may vary significantly between institutions and locations, and we expect authors to adhere to the NeurIPS Code of Ethics and the guidelines for their institution. 
        \item For initial submissions, do not include any information that would break anonymity (if applicable), such as the institution conducting the review.
    \end{itemize}

\item {\bf Declaration of LLM usage}
    \item[] Question: Does the paper describe the usage of LLMs if it is an important, original, or non-standard component of the core methods in this research? Note that if the LLM is used only for writing, editing, or formatting purposes and does not impact the core methodology, scientific rigorousness, or originality of the research, declaration is not required.
    \item[] Answer: \answerYes{} 
    \item[] Justification: This paper describes the usage of LLMs in our method.
    \begin{itemize}
        \item The answer NA means that the core method development in this research does not involve LLMs as any important, original, or non-standard components.
        \item Please refer to our LLM policy (\url{https://neurips.cc/Conferences/2025/LLM}) for what should or should not be described.
    \end{itemize}

\end{enumerate}
\clearpage

\end{document}